%% file: sample-manuscript.tex
\newcommand{\blue}[1]{#1}

\documentclass{acmart}

\AtBeginDocument{%
  \providecommand\BibTeX{{%
    \normalfont B\kern-0.5em{\scshape i\kern-0.25em b}\kern-0.8em\TeX}}}

\usepackage[utf8]{inputenc} 
\usepackage[T1]{fontenc}    
\usepackage{hyperref}       
\usepackage{url}            
\usepackage{booktabs}       
\usepackage{amsfonts}       
\usepackage{nicefrac}       
\usepackage{microtype}      
\usepackage{xcolor}         
\usepackage{algorithm}
\usepackage{algorithmic}

\usepackage{bm}
\usepackage{amsmath,amsthm}
\newtheorem{theorem}{Theorem}
\usepackage{graphicx}
\usepackage{tikz}
\usepackage{wrapfig}
\usepackage{subcaption}
\usepackage{multirow}

\setcopyright{acmlicensed}
\copyrightyear{2018}
\acmYear{2018}
\acmDOI{XXXXXXX.XXXXXXX}

\acmConference[Conference acronym 'XX]{Make sure to enter the correct
  conference title from your rights confirmation emai}{June 03--05,
  2018}{Woodstock, NY}
\acmISBN{978-1-4503-XXXX-X/18/06}

\begin{document}

\title{Differentially Private Low-Rank Adaptation of Large Language Model Using Federated Learning}

\author{Xiao-Yang Liu}
\email{xl2427@columbia.edu}
\affiliation{%
  \institution{Department of Computer Science, Rensselaer Polytechnic Institute. Department of Electrical Engineering, Columbia University}
  \streetaddress{530 W 120th Street}
  \city{New York}
  \state{NY}
  \country{USA}
  \postcode{10027}
}
\email{xl2427@columbia.edu}

\author{Rongyi Zhu}
\affiliation{%
  \institution{Department of Computer Science, University of Rochester}
  \streetaddress{229 Conant Road}
  \city{Rochester, NY}
  \country{USA}}
\email{rongyi.zhu@rochester.edu}

\author{Daochen Zha}
\affiliation{%
  \institution{Department of Computer Science, Rice University}
  \city{Houston}
  \country{USA}
\email{daochenzha@gmail.com}
}

\author{Jiechao Gao}
\affiliation{%
 \institution{Department of Computer Science, University of Virginia}
 \streetaddress{Rono-Hills}
 \city{Doimukh}
 \state{Arunachal Pradesh}
 \country{USA}}

\author{Shan Zhong}
\affiliation{%
  \institution{Department of Electrical Engineering, Columbia University}
  \city{New York}
  \state{New York}
  \country{USA}
  \email{sz2495@columbia.edu}}

\author{Matt White}
\affiliation{%
  \institution{Executive Director, PyTorch Foundation. GM of AI, Linux Foundation}
  \city{California}
  \state{California}
  \country{USA}
  \email{matt.white@berkeley.edu}}

\author{Meikang Qiu}
\affiliation{%
  \institution{School of Computer and Cyber Sciences, Augusta University }
  \streetaddress{Augusta}
  \city{Augusta}
  \state{Georgia}
  \country{USA}
  \postcode{78229}}
\email{ qiumeikang@yahoo.com}

\renewcommand{\shortauthors}{Trovato and Tobin, et al.}

\begin{abstract}

 The surge in interest and application of large language models (LLMs) has sparked a drive to fine-tune these models to suit specific applications, such as finance and medical science. However, concerns regarding data privacy have emerged, especially when multiple stakeholders aim to collaboratively enhance LLMs using sensitive data. In this scenario, federated learning becomes a natural choice, allowing decentralized fine-tuning without exposing raw data to central servers. Motivated by this, we investigate how data privacy can be ensured in LLM fine-tuning through practical federated learning approaches, enabling secure contributions from multiple parties to enhance LLMs. Yet, challenges arise: 1) despite avoiding raw data exposure, there is a risk of inferring sensitive information from model outputs, and 2) federated learning for LLMs incurs notable communication overhead. To address these challenges, this article introduces DP-LoRA, a novel federated learning algorithm tailored for LLMs. DP-LoRA preserves data privacy by employing a Gaussian mechanism that adds noise in weight updates, maintaining individual data privacy while facilitating collaborative model training. Moreover, DP-LoRA optimizes communication efficiency via low-rank adaptation, minimizing the transmission of updated weights during distributed training. The experimental results across medical, financial,  and general datasets using various LLMs demonstrate that DP-LoRA effectively ensures strict privacy constraints while minimizing communication overhead.

\end{abstract}

\keywords{Federated learning, differential privacy, domain-specific LLM, LLM privacy, medical LLM, financial LLM, efficient fine-tuning}

\received{20 February 2007}
\received[revised]{12 March 2009}
\received[accepted]{5 June 2009}

\maketitle

\section{Introduction}
\label{introduction}

Recent strides made in large language models (LLMs), like ChatGPT~\cite{ouyang2022training} and GPT-4~\cite{OpenAI2023GPT4TR}, have significantly propelled the field of natural language processing (NLP) forward. Their remarkable success has naturally sparked interest in domain-specific applications, such as finance~\cite{wu2023bloomberggpt, liu2023fingpt, yang2023fingpt,liu2023dynamic,xie2023pixiu}, medical science~\cite{singhal2023large}, law~\cite{nguyen2023brief}, biology~\cite{luo2022biogpt} and beyond. However, employing general-purpose LLMs directly in these specialized domains may result in subpar performance due to the inherent discrepancy between general and domain-specific texts~\cite{zha2023data-centric-survey,zha2023data-centric-perspectives}. Consequently, there arises an increasing demand to fine-tune LLMs for specific domains~\cite{hu2021lora,dodge2020fine,yu2021differentially,ding2023parameter}. Notably, OpenAI has released ChatGPT fine-tuning API\footnote{\url{https://platform.openai.com/docs/guides/fine-tuning}}, empowering users to tailor ChatGPT by fine-tuning it with their own datasets, effectively adapting it to their unique applications.


Despite the promise of LLM fine-tuning, there is a growing concern surrounding the potential leakage of private data while utilizing LLM~\cite{li2023privacy,sebastian2023privacy}, particularly within high-stakes domains. Consider, for instance, a scenario where multiple hospitals collaborate to enhance LLMs by fine-tuning them using their medical data. While this collaborative effort could significantly enhance model performance and offer substantial benefits to patients, hospitals may refrain from committing to this practice due to concerns about exposing sensitive patient information to other institutions and the public. In this scenario, federated learning strategies~\cite{li2020review,zhang2021survey} emerge as a natural choice, enabling distributed fine-tuning of the model with locally processed data, avoiding raw data exposure to servers. Therefore, we are motivated to study the following research question: \textbf{\emph{How can we guarantee data privacy in LLM fine-tuning with a practical federated learning approach, enabling each stakeholder to securely contribute data and derive collective advantages from an improved LLM?}}

However, accomplishing the aforementioned objective presents non-trivial challenges. \textbf{\emph{Firstly}}, despite federated learning ensuring no direct exposure to raw data, there remains a risk for attackers to potentially deduce training data from model outputs. This vulnerability arises because LLMs can accidentally disclose sensitive information through their responses~\cite{carlini2021extracting,kim2023propile,li2023privacy}. For instance, reports indicate that malicious adversaries could exploit the New Bing to link victims' personally identifiable information using partial information~\cite{li2023multi}. \textbf{\emph{Secondly}}, naively applying federated learning to fine-tune LLMs can trigger significant communication overhead. This is because model updates from diverse decentralized sources need iterative aggregation, and frequent transmission of these updates for a complex model like LLM is extremely expensive. Despite strategies to mitigate communication challenges in federated learning~\cite{konevcny2016federated,hamer2020fedboost,lan2023communication}, a notable trade-off persists: while these methods alleviate communication bottlenecks, they cannot ensure concrete guarantees in protecting the privacy of the training data.








To address the above challenges, in this paper, we introduce Differentially Private Low-Rank Adaptation (DP-LoRA), a communication-efficient federated learning algorithm designed for LLMs with privacy guarantees. \textbf{\emph{Ensuring data privacy}}, DP-LoRA employs a Gaussian mechanism, introducing random noise in weight updates to ensure minimal changes in publicly visible information when an individual in the dataset changes. This approach prevents any single sample from significantly impacting the output, thereby thwarting attackers' attempts to infer private information corresponding to specific individuals. We have substantiated our algorithm's adherence to the $(\epsilon,\delta)$-differential privacy property, a widely accepted criterion in differential privacy. \textbf{\emph{Addressing the communication overhead}}, DP-LoRA leverages the low-rank characteristic inherent in LLMs~\cite{hu2021lora} by incorporating a smaller set of new weights into the model, focusing fine-tuning solely on these parameters. Consequently, only a limited number of updated weights necessitate transmission during distributed training, resulting in substantial reductions in communication costs. In summary, this article contributes in the following ways:

\begin{itemize}
    \item We identify the need for preserving privacy in LLM fine-tuning and frame it as a federated learning paradigm, seeking to empower each stakeholder to securely contribute data and collectively benefit from enhancing the LLM.
    \item We present DP-LoRA, a novel federated learning algorithm specifically crafted for LLMs. DP-LoRA ensures differential privacy with privacy guarantees, while also reducing communication overhead via low-rank adaptation.
    \item Through experiments across medical, financial, and general datasets using four LLMs, we demonstrate the efficacy of DP-LoRA. The results consistently indicate its ability to maintain strict privacy constraints while minimizing communication overhead.

\end{itemize}

The rest of the paper is organized as follows: Section \ref{sec:relatedwork} provides a review of related work in LLM privacy, domain-specific LLM, parameter-efficient tuning for LLMs, federated learning, and differential privacy. Section \ref{sec:backgound} delves into the foundational principles of differential privacy, elucidating its crucial role in preserving data privacy during training. Moving forward, Section \ref{sec:method} details the proposed DP-LoRA algorithm for LLMs, which integrates low-rank adaptation with differential privacy to address communication costs while maintaining data privacy. Subsequently, Section \ref{sec:exp} presents the performance evaluation, analyzing and comparing the effectiveness of the proposed methods across general, medical, and finance datasets using various LLMs. Finally, Section \ref{sec:conclusion} concludes this paper.

\section{Related Work}
\label{sec:relatedwork}

\subsection{Privacy Issues of LLMs}

With the rise of LLMs, the issue of privacy has garnered significant attention, where it has been frequently shown that attackers can exploit model responses to infer sensitive information~\cite{carlini2021extracting,kim2023propile,li2023privacy,li2023multi,carlini2019secret}. This surge in concern has sparked a growing interest in preserving privacy within LLMs~\cite{shi2021selective,li2023privacy,yu2021differentially,anil2021large,hoory2021learning,li2021large}. However, current efforts tend to either focus solely on the pre-training phase of LLMs or introduce substantial communication overhead due to a large number of weight updates, significantly limiting the algorithm's practical application. Furthermore, these efforts primarily concentrate on general text data rather than domain-specific datasets. To tackle these limitations, we propose an efficient privacy-preserving fine-tuning algorithm for LLMs, aiming to ensure privacy with minimal communication overhead. We also showcase its effectiveness across medical, financial, and general datasets.

\subsection{From General-Purpose LLMs to Domain-specific LLMs}

The emergence of LLMs marks a significant shift in NLP~\cite{zhao2023survey}. Examples such as GPT-2/3/4~\cite{radford2019language,brown2020language,OpenAI2023GPT4TR}, GPT-NeoX~\cite{black2022gpt}, BERT~\cite{devlin2018bert}, PaLM~\cite{chowdhery2022palm}, BLOOM~\cite{scao2022bloom}, OPT~\cite{zhang2022opt}, and LLaMA~\cite{touvron2023llama} showcase exceptional language comprehension and generation abilities, excelling in various practical tasks. Recently, there has been a growing trend in crafting LLMs specialized for particular domains. In the following, we delve into recent LLMs efforts on medical and financial domains.

In the medical field, Med-PaLM~\cite{singhal2023large} emerges as a medical-specific LLM refined through instruction tuning using medical benchmarks, showcasing performance akin to clinicians. Additionally, BioGPT~\cite{luo2022biogpt}, a generative Transformer language model specifically pre-trained on expansive biomedical literature, demonstrates encouraging performance across diverse biomedical datasets. These initiatives across different domains have underscored the efficacy of domain-specific LLMs.

In the finance domain, the advent of FinLLMs starts with BloombergGPT~\cite{wu2023bloomberggpt}, a model trained on a blend of financial and general datasets. It demonstrates promise across crucial financial tasks like forecasting, sentiment analysis, and risk assessment. However, the closed-source nature of BloombergGPT's dataset poses obstacles to advancing FinLLMs, urging the exploration of cost-effective methods for domain adaptation. Recent endeavors, like FinGPT~\cite{liu2023fingpt}, aim to democratize access to vast internet-scale data for training financial LLMs by developing diverse data pipelines to acquire and refine high-quality training data. Additionally, Pixiu~\cite{xie2023pixiu} emerges as a tool for benchmarking LLMs in financial tasks, leveraging superior datasets to enhance the evaluation of model efficacy in this domain.

Despite the promise of domain-specific LLMs, the privacy issue remains prevalent. This concern becomes notably amplified in high-stakes domains such as finance and medical science. In this article, we aim to address this challenge by proposing efficient algorithms for preserving privacy during the fine-tuning process of domain-specific LLMs.

\subsection{Parameter-Efficient Tuning of LLMs}
The implementation of efficient fine-tuning methods for LLMs has garnered significant attention~\cite{hu2021lora,dettmers2023qlora,lester2021power,sung2022lst,liu2023winner,zaken2021bitfit,karimi2021compacter}. Notably, some widely recognized approaches include the utilization of adapters~\cite{houlsby2019parameter,karimi2021compacter}, which embed a compact module within transformer blocks, allowing selective updates while keeping other parameters fixed. Similarly, the Low-Rank Adaptation (LoRA) technique~\cite{hu2021lora} integrates trainable rank decomposition matrices into transformer blocks, facilitating updates solely to these matrices while preserving the remaining parameters. However, these parameter-efficient tuning methods do not consider data privacy. In this article, we propose leveraging federated learning and differential privacy to enable secure fine-tuning of LLMs.

\subsection{Federated Learning and Differential Privacy}

Federated Learning, a recently proposed training approach, focuses on safeguarding dataset privacy during distributed training. Early implementations involved employing model averaging to minimize communication overhead \cite{mcmahan2017communication}. Building upon this foundation, recent advancements have introduced various algorithms, including variance reduction methods \cite{das2022faster} and penalty-based approaches \cite{li2020federated}, in both homogeneous \cite{woodworth2020local} and heterogeneous \cite{karimireddy2020scaffold} settings. Notably, the integration of differential privacy into federated learning has gained traction, offering a theoretical framework to govern privacy boundaries and manage the tradeoff between privacy and convergence \cite{wei2020federated,konevcny2016federated,hamer2020fedboost,lan2023communication}. However, the utilization of differential privacy methods often results in substantial communication costs, particularly when applied to LLMs. In this article, we tackle this challenge through low-rank adaptation. The following section will provide more background of differential privacy.

\section{Background of Differential Privacy}
\label{sec:backgound}

 \textbf{Deep neural networks}.
For simplicity, we start with a single-layer network, which takes an input vector $\bm{x} \in \mathbb{R}^{m}$ and transforms it into an output vector $\bm{\hat{y}} \in \mathbb{R}^{n}$ via a weight matrix $\bm{W} \in \mathbb{R}^{n \times m}$,
\begin{equation}
    \bm{\hat{y}} = \bm{W} \cdot \bm{x} + \bm{b},
\end{equation}
where $\bm{b} \in \mathbb{R}^{n}$ is a bias vector. Let $\bm{\theta}$ denote the trainable parameters, i.e., $\bm{\theta} = \{\bm{W}, \bm{b}\}$. A loss function $\mathcal{L}(\bm{\theta}; \bm{x}, \bm{y})$ is used to evaluate the difference between $\bm{\hat{y}}$ and $\bm{y}$, where $\bm{y}$ is the ground-truth label.

    

Assuming the training dataset $D$ has $N$ data pairs. During the training process, each iteration uses a batch to update the parameter $\bm{\theta}$, where each batch contains $B$ samples $\{ (\bm{x}^b, \bm{y}^b) \}_{b=1}^B$. The parameter $\bm{\theta}$ is updated by the gradient descent method as follows
\begin{align}
    \bm{g} &= \frac{1}{B} \nabla_{\bm{\theta}} \sum\limits_{b=1}^{B} \mathcal{L}(\bm{\theta}; \bm{x}^b, \bm{y}^b), \\
    \bm{\theta} &= \bm{\theta} - \gamma \cdot \bm{g},
\label{eq:sgd}   
\end{align}
where $\gamma$ is the learning rate. The training process executes for $T$ iterations.

\textbf{Federated learning}.
We consider distributed training of a deep neural network with parameter $\bm{\theta}$. There are $K$ nodes participating in distributed training. Each node $k$ performs the gradient descent using its private dataset $D_k$, which contains $N_k$ samples. Thus, for node $k$, we have the sampling probability $q_k = B/N_k$. There are $N$ samples in total, where $N = \sum_{k=1}^{K}N_k$. In particular, one distributed training iteration consists of the following three:
\begin{enumerate}
    \item[1)] A server broadcasts a global model $\bm{\theta}$ to all $K$ nodes.
    \item[2)] Node $k$ initializes $\bm{\theta}_k$ by $\bm{\theta}$ and updates $\bm{\theta}_k$ by (\ref{eq:sgd}), using a batch of $B$ samples from $D_k$.
    \item[3)] The server receives $\{\bm{\theta}_1,...,\bm{\theta}_K\}$ from all $K$ nodes and updates $\bm{\theta}$ by
    \begin{equation}
        \bm{\theta} = 
        \sum\limits_{k=1}^{K} \rho_k \cdot \bm{\theta}_k,
    \label{eq:avg}
    \end{equation}
\end{enumerate}
where the $\rho_k$ is the weight factor, i.e., $\sum_{i=1}^{K}\rho_k = 1$. The training process executes for $T$ iterations. The training process executes for $T$ iterations.

\textbf{Differential privacy}.
However, the distributed training comes with a privacy issue \cite{carlini2021extracting}. Some attackers surveil the communication between the server and nodes and infer the information in dataset $D_k$. We need to provide a privacy guarantee for neural network training. Differential privacy considers the divergence of two distributions. There are multiple options to define the distance between distributions, and KL divergence is a popular choice. On the other hand, a $(\epsilon, \delta)$-differential private mechanism $\mathcal{M}$ is defined as 
\begin{equation}\label{def:dp}
    \text{Pr}[\mathcal{M}(d) = o ] \leq e^\epsilon ~ \text{Pr}[\mathcal{M}(d') = o ] + \delta,
\end{equation}
where $d$ and $d'$ are two adjacent inputs and $o$ stands for the output of mechanism $\mathcal{M}$. The divergence is bounded by $\epsilon$, and a relaxation of $\delta$ is allowed. Thus, we introduce randomness into the training process.

DP-SGD in \cite{abadi2016deep} proposes to add Gaussian noise into the gradient and provides a privacy guarantee for network training. For each iteration, the gradient $\bm{g}$ is replaced by: 
\begin{equation}
    \widetilde{\bm{g}} = \frac{\bm{g}}{\max(1, \frac{\bm{||\bm{g}||_2}}{C})} + \mathcal{N}(0, \sigma^2C^2\bm{I}),
\label{eq:gradient}
\end{equation}
where $C$ is the clipping bound and $ \mathcal{N}(0, \sigma^2C^2\bm{I}) $ is the Gaussian distribution with mean 0 and standard deviation $\sigma^2C^2\bm{I}$. Under the Gaussian mechanism, each iteration is differentially private. 

The training process can be seen as a composition of $T$ differentially private algorithms. 

\begin{theorem} \label{thm:seq} (Sequential composition \cite{mcsherry2009privacy})
Let $\mathcal{M}_t $ each provides $(\epsilon_t, \delta_t)$-differential privacy, $t = 1,2,...,T$. The sequential composition of $\mathcal{M}_t $ provides $(\epsilon_t^{'}, \delta_t^{'})$-differential privacy, where the $\epsilon_t^{'} = \sum_t\epsilon_t$ and $\delta_t^{'} = \sum_t \delta_t$. 
\end{theorem}

In Theorem \ref{thm:sigma}, we can be shown that the training process essentially follows the differential privacy mechanism, and by incorporating moment counting, the noise has a scale of $\sigma$. 

\begin{theorem} \label{thm:sigma}
 \cite{abadi2016deep}  There exists constant $c_1$ and $c_2$ so that with the sampling probability $q = B/N$ over $T$ iterations, for any $\epsilon < c_1 q^2 T$, the network training process using DP-SGD method is $(\epsilon, \delta)$-differentially private for a given $\delta > 0$ as long as 
    \begin{equation}
    \sigma \geq  c_2 \frac{q\sqrt{T\log(1/\delta)}}{\epsilon}.
    \label{eq:delta}
    \end{equation}
\label{the1}
\end{theorem}





\begin{figure}
    \centering
    \includegraphics[width=\linewidth]{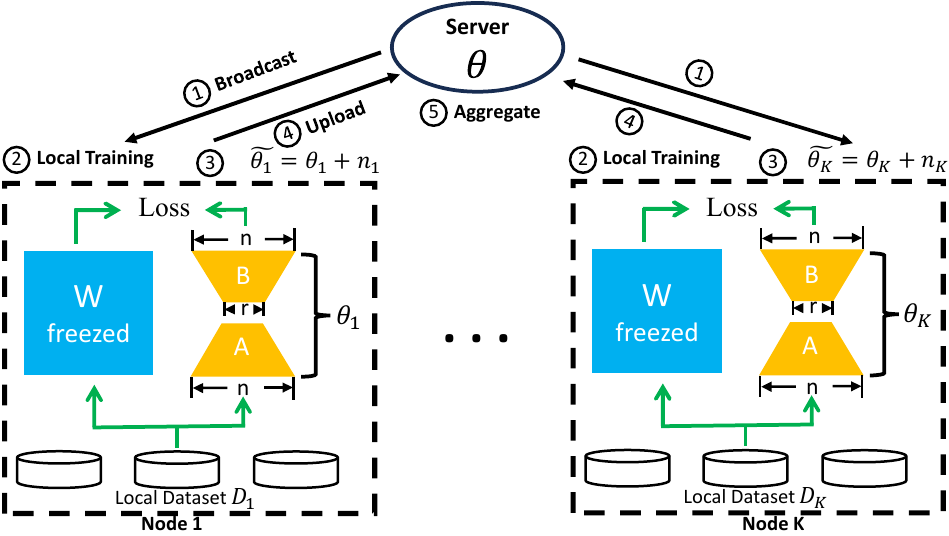}
    \caption{Differential private low-rank adaptation (DP-LoRA) algorithm for LLMs using federated learning. Each node communicates with the server in updating $\bm{\theta}$. }
   \label{fig:fl}
\end{figure}

\section{Differentially Private Low-Rank Adaptation (DP-LoRA) of LLMs}
\label{sec:method}



\blue{We propose a differentially private algorithm for a large language model with a federated learning scenario. First, we elaborate on the privacy guarantee in a non-balance federated learning of our methods. Second, our method is able to mitigate the communication overhead in LLM weight transmission.} 

\subsection{\blue{Motivation}}
\label{sec:motivation}
Multiple institutes collaborate in training a large language model using their local datasets. Current advancements in large language models have led to an increased focus on open-source models by multiple institutions, such as the Linux Foundation, Hugging Face, and AI4Finance Foundation. These organizations collaborate in training large language models in vertical domains, such as medicine and finance. Each institute in the organization processes a local dataset for collaborative training. For instance, banks cooperate in training financial models on transaction strategies and risk management with their financial data. Hospitals collaborate to develop models for diagnosing and treating medical conditions with medical data. These collaborations often involve training existing large language models with data from the specific vertical domain. Each participant contributes their data in this collaboration framework.


Each participating institution usually possesses sensitive, private data. Due to the constraints of privacy laws, these data cannot be shared among collaborating entities. While federated learning has been widely studied for these scenarios, recent studies have discovered that the federated learning approach is not safe enough to protect private data in large language models \cite{carlini2021extracting}. In the meantime, the communication overhead becomes imminent as the model size becomes very large.

In this work, we proposed DP-LoRA, a cooperation scheme for large language model federated learning. Firstly, we prove that DP-LORA is a privacy-preserving mechanism that provides a privacy guarantee for sensitive training data. Secondly, we introduce the low-rank adaptation into federated learning to mitigate the communication overhead. Thirdly, we elaborate a weighted federate learning scheme that corresponds to varying data quantities and qualities.

\subsection{DP-LoRA Algorithm}
\label{sec:alg}


Our method is shown in Fig.~\ref{fig:fl}. There are $K$ nodes, where every node $k$ holds a local dataset $D_k$ and a large language model. For the dataset, $D_k$ has $N_k$ samples and is not shared with the server or other nodes. We use $N$ to denote the total number of samples, where $N = \sum\limits_{k=1}^{K}N_k$. For the large language model, every node has the same model weight $\bm{\theta}$ in the beginning. $\bm{\theta}$ includes the original LLM weight $\bm{W}$ and the low-rank adaptation weight $\bm{A}$ and $\bm{B}$. Each node $k$ freezes the original model weight $\bm{W}$ and fine-tunes the $\bm{A}_k$ and $\bm{B}_k$. 



\blue{Given the noise scale $\sigma$, the clipping norm $C$, the number of iterations $T$, the learning rate $\gamma$, and the batch size $B$, we propose a differential private low-rank adaptation (DP-LoRA) algorithm for federated learning, while the pseudocode are summarized in Alg. \ref{alg:local}.}

\begin{itemize}
    \item[1] \textbf{Broadcast}: in line 4 of Alg. \ref{alg:local}, the server sends the global model parameters $\bm{\theta}$ to all participating nodes. This is the starting point for training in each round.
    \item[2] \textbf{Local Training}: \blue{ In line 11-14 of Alg. \ref{alg:local}, each node $k$ samples $B$ samples from $D_k$ and uses the clipped gradient to update the local weight $\bm{A}_k$ and $\bm{B}_k$.}
    \item[3] \textbf{Add Noise}: In line 16 of Alg. \ref{alg:local}, a Gaussian noise $n_k \sim \mathcal{N}(0, \sigma^2)$ is add to the $\bm{A}_k$ and $\bm{B}_k$ simultaneously. We denote the noise weight as $\widetilde{\bm{A}_k} = \bm{A}_k + n_k$ and $\widetilde{\bm{B}_k} = \bm{B}_k + n_k$. 
    \item[4] \textbf{Upload}: In line 5 of Alg. \ref{alg:local}, every node $k$ uploads $\widetilde{\bm{A}_k}$ and $\widetilde{\bm{B}_k}$ to the server.
    \item[5] \textbf{Aggregate}: \blue{In line 6 of Alg.\ref{alg:local}, the server aggregates the $\widetilde{\bm{A}_k}$ and $\widetilde{\bm{B}_k}$ from all nodes and takes the weighted average $\sum\limits_{k=1}^{K} \rho_k \widetilde{\bm{A}_k}$ and $\sum\limits_{k=1}^{K} \rho_k \widetilde{\bm{B}_k}$ to updates $\bm{A}$ and $\bm{B}$ as follows
    \begin{equation}
    \label{eq:aggreagate}
    \widetilde{\bm{A}} = \sum\limits_{k=1}^{K}(\bm{A}_k + n_k) =\sum\limits_{k=1}^{K} \rho_k \bm{A}_k + \sum\limits_{k=1}^{K} \rho_k n_k, \quad \widetilde{\bm{B}} = \sum\limits_{k=1}^{K}(\bm{B}_k + n_k) =\sum\limits_{k=1}^{K} \rho_k \bm{B}_k + \sum\limits_{k=1}^{K} \rho_k n_k
\end{equation}
where $\rho_k = N_k/N$.}
\end{itemize}
We elaborate the process in lines 10-18 of Alg. \ref{alg:local}. \blue{ First, calculate the gradients of the loss function $\mathcal{L}$ for the parameters $\bm{A}_k$ and $\bm{B}_k$, denoted as $\bm{g}_{\bm{A}_k}$, $\bm{g}_{\bm{B}_k}$. Second, perform gradient clipping by scaling down the gradients if their L2 norm is greater than $C$, $\bm{g}_{\bm{A}_k} \leftarrow \bm{g}_{\bm{A}_k} / \max(1, \frac{\| \bm{g}_{\bm{A}_k} \|_2}{C})$ and $\bm{g}_{\bm{B}_k} \leftarrow \bm{g}_{\bm{B}_k} / \max(1, \frac{\| \bm{g}_{\bm{B}_k} \|_2}{C})$. Third, introduce Gaussian noise to the clipped gradients to ensure differential privacy, $\bm{g}_{\bm{A}_k} \leftarrow  \bm{g}_{\bm{A}_k} + \mathcal{N}(0, \sigma^2 C^2 \bm{I})$ and $\bm{g}_{\bm{B}_k} \leftarrow  \bm{g}_{\bm{B}_k} + \mathcal{N}(0, \sigma^2 C^2 \bm{I})$, $k = 1,2,...,K$. Fourth, update the local model parameters via gradient descent, $\widetilde{\bm{A}_k} \leftarrow \bm{A}_k - \gamma \bm{g}_{\bm{A}_k}$ and $\widetilde{\bm{B}_k} \leftarrow \bm{B}_k - \gamma \bm{g}_{\bm{B}_k}$, where $\bm{g}_k$ is the noised and clipped gradient. }

\subsection{Guarantee of Differential Privacy}
\label{sec:dp-lora}


\blue{The differentially private low-rank adaptation (DP-LoRA) we introduced is a $(\epsilon, \delta)$ differential private mechanism. We prove that in Theorem \ref{thm:sigma}. }

\begin{algorithm}[t]
    \renewcommand{\algorithmicrequire}{\textbf{Input:}}
    \renewcommand{\algorithmicensure}{\textbf{Output:}}
    \caption{DP-LoRA Algorithm}
    \label{alg:1}
    \begin{algorithmic}[1]
        \STATE \textbf{Input}: number of nodes $K$, dataset $ \{ D_k \}_{k=1}^{K}$, noise scale $\sigma$, clipping norm $C$, iterations $T$, learning rate $\gamma$, batch size $B$, network parameter $\bm{\theta}$.
        
        \STATE Initialize $\bm{\theta}$
        \FOR{$t=0$ to $T$}
            \STATE Send $\bm{\theta}$ to all $K$ nodes independently
            
            \STATE $\bm{A}_k, \bm{B}_k$ $\gets$ Nodeupdate($D_k$, C, B, $\sigma$, $\gamma$),  ~~$k=1,2,...,K$
            \STATE $\bm{A} \gets  \sum\limits_{k=1}^{K} \rho_k \bm{A}_k$, \quad $\bm{B} \gets  \sum\limits_{k=1}^{K} \rho_k \bm{B}_k$
        \ENDFOR
        \STATE \textbf{Output}: $\bm{A}$, $\bm{B}$
        \STATE 
        \STATE \textbf{function} Nodeupdate($D_k$, C, B, $\sigma$, $\gamma$)
        \STATE \textbf{Input}: dataset $D_k$, clipping norm $C$, learning rate $\lambda$, batch size $B$ and noise scale $\sigma$.
        \STATE sample $(\bm{x}, \bm{y})$ from $D_k$ with batch size $B$
        \STATE $\bm{g}_{\bm{A}_k} \gets \nabla_{\bm{A}_k} \mathcal{L}(\bm{g}_{\bm{A}_k}; \bm{x}, \bm{y})$, \quad $\bm{g}_{\bm{B}_k} \gets \nabla_{\bm{B}_k} \mathcal{L}(\bm{g}_{\bm{B}_k}; \bm{x}, \bm{y})$, 
        \STATE $\bm{g}_{\bm{A}_k} \gets \bm{g}_{\bm{A}_k} / \max(1, \frac{\Vert \bm{g}_{\bm{A}_k} \Vert_2}{C})$, \quad $\bm{g}_{\bm{B}_k} \gets \bm{g}_{\bm{B}_k} / \max(1, \frac{\Vert \bm{g}_{\bm{B}_k} \Vert_2}{C})$ 
        \STATE $\bm{g}_{\bm{A}_k} \gets \bm{g}_{\bm{A}_k} + \mathcal{N}(0, \sigma^2C^2\bm{I})$, \quad $\bm{g}_{\bm{B}_k} \gets \bm{g}_{\bm{B}_k} + \mathcal{N}(0, \sigma^2C^2\bm{I})$  \COMMENT{$\sigma \geq  c_2 \frac{q\sqrt{T\log(1/\delta)}}{\widebar{\rho}\epsilon}$}
        \STATE $\bm{A}_{k} \gets \bm{A}_{k} - \gamma  \bm{g}_{\bm{A}_k}$, \quad $\bm{B}_{k} \gets \bm{B}_{k} - \gamma  \bm{g}_{\bm{B}_k}$
        \STATE \textbf{Output} $\bm{A}_k$, $\bm{B}_k$
        \STATE \textbf{end function}
    \end{algorithmic}
    \label{alg:local}
\end{algorithm}







\blue{For a single node $k$, we add Gaussian noise in the gradient when we update $\bm{A}_k$ and $\bm{B}_k$. The probability density function for Gaussian noise is $ \mathcal{N}(0, \sigma^2C^2\bm{I}) $, where $\bm{I}$ is an identity matrix and has the same shape with $\bm{A}_k$ and $\bm{B}_k$. The clipping bound $C$ is used to ensure the gradient norm $|| \bm{g}_{\bm{A}_k} || < C$ and $|| \bm{g}_{\bm{B}_k} || < C$. The noise scale $\sigma$ provides the privacy guarantee for each iteration, which should follow $\sigma \geq \sqrt{2\ln{(1.25/\delta)}}/\epsilon$. We use $\widetilde{\bm{A}_k}$ and $\widetilde{\bm{B}_k}$ to denote the noise weight from the node $k$.}


\blue{For the server, it aggregates all $\widetilde{\bm{\theta}_k}$ uploaded from the server by Eq. (8). 
we prove that the differential private property holds for the proposed algorithm, which guarantees the privacy for the training dataset $D_k$.}

\blue{
\begin{theorem} \label{thm:sigma} There exists positive constants $c_1 > 0 $ and $c_2 >0$, the sampling probability $q = B/N$, the batch size $B$ and dataset size $N$, for any $\epsilon < c_1 q^2 T$, DP-LoRA with $K$ nodes in Alg. \ref{alg:local} over $T$ iterations is $(\epsilon, \delta)$-differentially private for a given $\delta > 0$ such that
    \begin{equation}
    \sigma \geq  c_2 \frac{q\sqrt{T\log(1/\delta)}}{\widebar{\rho}\epsilon},
    \label{eq:delta}
    \end{equation}
\label{the1}
where $\widebar{\rho} = \sqrt{\rho^2_1 + \rho^2_2 + ... + \rho^2_k}$ and $\sum\limits_{k=1}^{K}\rho_k = 1$
\end{theorem}. 
In the above theorem, we assume that all the node adopts the same sample probability $q$. We detailed the proof in our appendix.}



\subsection{Reducing Communication Overhead in Federate Learning}
\label{method:fl}
In this section, we illustrate how the proposed DP-LoRA would reduce the communication overhead. Typically, a language model with wight $\bm{W}$ is composed by $L$ weight matrices $\bm{W}^l \in \mathbb{R}^{n \times n}, l = 1,2,..., L$. Here, we consider the linear layers, and the input of output shares the same dimension $n$. These weight matrices have a low “intrinsic dimension" \cite{aghajanyan2020intrinsic}. We decompose the matrix $\bm{W}^l$ into $\bm{A}^l \in \mathbb{R}^{n \times r}$ and $ \bm{B}^l \in \mathbb{R}^{r \times n}$, follow by low-rank adaptation method \cite{hu2021lora}. Note that the rank $r \ll n $. Thus, we have $\bm{\theta} = \{\bm{A}^1, \bm{B}^1; \bm{A}^2, \bm{B}^2;...; \bm{A}^L, \bm{B}^L \}$. 


The weight $\bm{\theta}$ is composed by $\bm{A}$ and $\bm{B}$. Federated learning involves collecting local model weights for $T$ iterations and redistributing the aggregated weights $\bm{\theta}$ back to each participating node. It is assumed that all nodes utilize Algorithm~\ref{alg:local} for training and share identical hyper-parameters. The system comprises $K$ nodes in total. Each node possesses its private dataset $D_k$, where $k = 1,2,\ldots,K$, and maintains local model weights $\bm{\theta}_k$, for $k = 1,2,\ldots,K$. In Algorithm~\ref{alg:local}, each node $k$ refines its $\bm{\theta}_k$ using the corresponding dataset $D_k$. A central server maintains the global model weights $\bm{\theta}$.

We assess the reduction in communication overhead achieved through the low-rank adaptation method. Assuming that the federated learning process spans $T$ rounds in a noiseless and ideal network environment, the communication overhead can be quantified using specific parameters. The proposed method provides the reduction in two steps. First, we freeze most of the model weight. We fine-tune and  transfer the weight in self-attention layers. For example, a Llama-7B is composed of $32$ transformer layers and each transformer consists $32$ attention heads, which have $4096$ dimensions. Thus, each self-attention block contains $4096\times4096\times3 = 50,331,648$ parameters, and $32$ transformers contain $1,610,612,736$ parameters in total. Therefore, we reduce the training parameter from $6,738,411,520$ to $1,610,612,736$. Second, we decomposed the self-attention parameters by low-rank adaptation methods. We use a $\bm{W}^l$ to denote the parameters of one attention head. Each $\bm{W}^l \in \mathbb{R}^{ n \times n}$ is replace by a $\bm{A}^l \in \mathbb{R}^{n \times r}$ and $\bm{B}^l \in \mathbb{R}^{n \times r}$. Thus, for one attention head with $4096$ dimension, we reduce the parameter from $16,777,216$ to $2,097,152$ if we choose $r = 256$. 

In general, the communication overhead is determined by the number of nodes $K$, the training epoch $T$, and the transmission parameters $\bm{\theta}$. The total communication overhead is given by $T \times K \times L \times r \times 2n$. Take training Llama-7B with 5 nodes in 50 iterations for example, the original communication overhead is $2 \times 5 \times 50 \times 6.7B$ and the communication overhead for DP-lora is $2 \times 5 \times 50 \times 4096 \times 256$. 


\section{Performance Evaluation}
\label{sec:exp}


\subsection{Tasks and Datasets}
\label{sec:exp1}

To mimic the specific implementation scenario, we chose three datasets to test our algorithm. To mimic the protection of highly sensitive data, we choose the financial and medical data which always contains sensitive information. We also test our algorithm on a normal dataset to provide a comparison with other methods. \blue{In the following, we primarily present the results on the medical datasets, with the financial datasets provided in the appendix.}

\noindent \textbf{SlimPajama}: open-source dataset for training large language models. Slimpajama is created from RedPajama \cite{together2023redpajama} by filtering out low-quality data and duplicates. SlimPajama contains 627B tokens. The data in Slimpajama mainly comes from a wide array of online sources. (1). Arxiv: ArXiv contains preprints in fields like physics, mathematics, and computer science, offering early access to scholarly research articles before peer review. (2). Github: GitHub contains tools for code hosting, collaboration, version control, and project management for software development. (3). Wikipedia: Wikipedia contains a vast range of information, including articles on topics like history, science, geography, technology, culture, biographies, and much more, all written and continually updated by volunteer contributors from around the world.

\noindent \textbf{Medical dataset}:  dataset contains 0.26 million English consultations between patients and doctors. The total number of utterances is 0.51 million. Each consultation consists of two parts: (1) a description of the patient’s medical conditions; (2) a conversation between the patient and the doctor. The data is crawled from "iclinic.com3" and "healthcaremagic.com", which are two online platforms of healthcare services, including symptom self-checker, video consultation, online chat with doctors, etc. The consultations cover 51 categories of communities including diabetes, elderly problems, pain management, etc., and 96 specialties including andrology, cardiology, nephrology, pharmacology, etc. The consultations were conducted from 2008 to 2020.

\noindent \textbf{Finance dataset}: We collect extensive financial data from various sources. (1). Financial news: Websites such as Reuters, CNBC, and Yahoo Finance, among others, are rich sources of financial news and market updates. (2). Social media: Platforms such as Twitter, Facebook, Reddit, Weibo, and others, offer a wealth of information in terms of public sentiment. (3). Filings: Websites of financial regulatory authorities, such as the SEC in the United States, offer access to company filings. (4). Trends: Websites like Seeking Alpha, Google Trends, and other finance-focused blogs and forums provide access to analysts’ opinions, and market predictions. (5). Academic datasets:  Research-based datasets that offer curated and verified information for sophisticated financial analysis. The dataset comprises 9,540 samples for training, each annotated with one of three labels: Bearish, Bullish, or Neutral.

\blue{Note that the aforementioned datasets often lack truly sensitive information due to their open-source nature. Our algorithm does not differentiate between sensitive and non-sensitive portions of the data; instead, it focuses on protecting the local dataset as a whole. Users can readily apply our algorithm to data they deem sensitive.}

\input{tabs/dataset_statics}

\subsection{Experiments Setup}
\textbf{Evaluation Task}: We assess our methods using test sets across different fields, aligned with the training data.

For general tasks, we use three datasets: BoolQ, a question-answering dataset with 16K examples, each containing a passage, a related yes/no question, and an answer. PIQA, focused on physical commonsense reasoning, comprising 21K questions with two solution options, where the model selects the most appropriate. WinoGrande, a commonsense reasoning dataset with 44K sentences, each with a blank to fill using one of two provided options.

In the financial domain, we use FPB, with 4.8K financial news sentences categorized by sentiment (positive, negative, neutral). FiQA SA, comprising 17K microblog headlines and financial news sentences, is classified by sentiment. TFNS, a collection of 11.9K finance-related tweets, is used for sentiment classification.

For medical tasks, our datasets include MedQuAD, which offers 47,457 medical question-answer pairs from 12 NIH websites, covering 37 question types. LiveQA Test, featuring questions across 26 types and five categories, each with subquestions, a focus, and validated reference answers from trusted resources. MEDIQA-Ans, containing 156 medical questions.


\noindent \textbf{Models}: To provide a comprehensive understanding of our algorithm,  we test our algorithm based on different language models. (1) GPT-2\cite{radford2019language}: We chose the basic version of GPT-2 in our experiments. It has 1.5 billion parameters and it is composed of 12 layers and each layer has 768 dimensions. The model is released on \cite{radford2019language}. (2) Bert \cite{devlin2018bert}: We chose the Bert base version to conduct our experiments. The Bert base uses 12 layers and 12 attention heads and has a total number of Parameters of 110M. The pre-trained weight can be found in the Link from \cite{devlin2018bert}. (3) ChatGLM-6b \cite{zeng2022glm}: ChatGLM2-6B has 6.2 billion parameters. It contains 28 layers with dimensions of 4096 and heads of 32.  (4) Llama2-7B \cite{touvron2023llama}: Llama2-7B contains 6.7B parameters. It contains 32 layers with dimensions of $4096$ and attheads of $32$. The initial Llama-2 weights are downloaded from the given link \cite{touvron2023llama}. 

\noindent \textbf{Experiments settings}: For comparison, we choose three federated learning algorithms as our baseline. (1)Vanilla FL (2)T-FedAvg~\cite{xu2020ternary} (3)CMFL \cite{luping2019cmfl}. We train the baselines and the proposed method in the same settings. We split the dataset evenly and stored it locally among all $K$ nodes. We adopt an average for aggregating local model weights and we set all $\rho_k $ in Equation \ref{eq:avg} to $ \frac{1}{K}$. We assume that the communication channels are noiseless and ideal.  Each node $k$ uploads its $\bm{\theta}_k$ to the server in every iteration. We choose $K$ as $5$ and we set the maximum iteration $T$ as 50. For the training, we follow the settings in Table. \ref{tab:hyper}. 

\input{tabs/hyperparamter}

\subsection{Results on Medical Tasks}

In our experiments, we evaluated the performance of several models in the context of differential privacy for medical tasks. We investigated the impact of varying the parameters $\epsilon$ and $\delta$ for the language model fine-tuning on the medical datasets. In all experiments, the compression rank was fixed at $512$ and the $\delta$ is set to $1e-5$ when we varying the $\epsilon$ and the $\epsilon$ is set to $2$. 

One notable trend is the general decrease in performance across all models with stricter privacy settings (lower $\epsilon$ and lower $\delta$ values). For instance, Llama-7B's performance on LiveQA drops from $69.4$ at the original setting to $55.9$ and $49.3$ when $\epsilon$ is reduced to $2$ and $\delta$ to $1e-05$, respectively (Table \ref{tab:epsilon_med} and Table \ref{tab:delta_med}). This trend indicates a trade-off between privacy and utilities.

\input{tabs/epsilon_med}

In contrast, some models like ChatGLM-6B show a more resilient performance under varying $\epsilon$ values. For example, its performance on LiveQA only marginally decreases from $71.9$ to $69.8$ when $\epsilon$ is increased from the original to 10 (Table \ref{tab:epsilon_med}). This suggests that certain models might be better suited for privacy-sensitive applications.


\input{tabs/delta_med}

We can observe a significant trade-off between privacy and utility in medical tasks using models like GPT-2, Bert, ChatGLM-6B, and Llama-7B. Our data clearly shows that stricter privacy settings (lower $\epsilon$ and $\delta$ values) generally correlate with reduced performance across tasks.

\input{tabs/compression}

\subsection{Reduction in Communication Overhead}

 We test the performance of the global model and communication overhead in federated learning. We choose the rank $r$ in a decreasing manner. We use model parameter amount to indicate communication overhead. The local models are trained with Alg. \ref{alg:local}. We set the privacy parameters $\epsilon = 8$ and $\delta=10e-5$. 

Across all models, we observed a consistent trend: as the rank $r$ decreased, the number of parameters and consequently the communication overhead also decreased. This trend indicates a direct correlation between rank reduction and communication efficiency. However, there is a trade-off observed in model performance on general tasks for the differential privacy training.

Llama-7B (Table \ref{tab:compression}), one of the most popular large language models, maintained a relatively stable performance (e.g., BoolQ at 76.3 to 39.1, WinoGrande 70.0 to 67.8 ) despite significant parameter reduction (compression ratio from 7.31 to 36.27). In contrast, ChatGLM-6B (Table \ref{tab:compression}) showed a more pronounced performance degradation (BoolQ dropping from 69.5 to 50.1, WinoGrande from 71.3 to 46.3) with a notable decrease in parameters (compression ratio moving from 7.6 to 47.4). These results highlight the variability in how different models respond to parameter reduction and compression, underscoring the challenge of balancing communication efficiency with performance in federated learning environments.

\section{Conclusion}
\label{sec:conclusion}

In this paper, we have introduced DP-LoRA, an innovative approach to preserve data privacy in the low-rank adaptation of LLMs. DP-LoRA ensures differential privacy while significantly reducing communication overhead. By introducing noise into gradients to achieve ($\epsilon$, $\delta$)-differential privacy, we enhance communication efficiency. Empirical validation through experiments demonstrates the efficacy of our approach, rendering LLMs more secure for domain-specific fine-tuning. Looking ahead, our future research endeavors will explore refining the error bounds for differential privacy, particularly in the context of fine-tuning LLMs where the existing bounds may be less stringent. Additionally, we aim to broaden our experimental scope to encompass a wider array of domain-specific datasets, further validating the applicability and robustness of DP-LoRA. \blue{Finally, we plan to conduct more experimental analysis to address the scenarios where different institutions possess their own unique datasets that do not conform to the same
distribution as others.}

\bibliographystyle{unsrtnat}
\bibliography{sample-base}

\section{Appendix}
\label{sec:appendix}
%





\subsection{\blue{Detailed Proof}}
\label{sec:appendix_proof}
\blue{As stated in the main part, our algorithm satisfies $(\epsilon, \delta)$-differential privacy as follows. }

\blue{
\begin{theorem} \label{thm:sigma} There exists positive constants $c_1 > 0 $ and $c_2 >0$, the sampling probability $q = B/N$, for any $\epsilon < c_1 q^2 T$, the training process using the DP-LORA method over $T$ iterations is $(\epsilon, \delta)$-differentially private for a given $\delta > 0$ such that
    \begin{equation}
    \sigma \geq  c_2 \frac{q\sqrt{T\log(1/\delta)}}{\widebar{\rho}\epsilon}.
    \label{eq:delta}
    \end{equation}
\label{the1}
\end{theorem}  }

\begin{proof}




\blue{
There are $k$ nodes participating in the training of LLMs. The server collects noised weight $\widetilde{\bm{\theta}_k}$ from all nodes and aggregates all weights in each iteration. For every node, $\widetilde{\bm{\theta}_k}$ is obtained following the steps in Alg. \ref{alg:local}. To prove the overall differential privacy property, we look at the mechanism in the server, which states as (\ref{eq:aggreagate}).}

\blue{
Let us define a random variable, \textit{privacy loss}, as follows:
\begin{equation}
    c(o) = \ln \left(\frac{\Pr[\mathcal{M}(d) = o] }{\Pr [\mathcal{M}(d') = o]} \right)
\end{equation}
The $(\epsilon, \delta)$-differential privacy in (\ref{def:dp}) is mathematically as follows
\begin{equation}
    \Pr\left[\ln\left(\frac{\Pr[\mathcal{M}(d) = o]}{\Pr[\mathcal{M}(d') = o]} \right) \geq \epsilon \right] \leq \delta,
\end{equation}
where the divergence is bounded by $\epsilon$ and a relaxation of probability $\delta$ is allowed.}




\blue{Next, we consider the log of the moment generating function evaluated at the value $\lambda$:
\begin{equation}
\alpha_\mathcal{M}(\lambda; \widetilde{\bm{\theta}}, d, d') \triangleq \log
\mathbb{E}_{o \sim \mathcal{M}(\widetilde{\bm{\theta}}, d)}
[\exp(\lambda c(o) ]
\end{equation}
\indent To bound all possible $\alpha_{\mathcal{M}}(\lambda; \bm{\theta}, d, d^{'})$, $\alpha_{\mathcal{M}}(\lambda)$ is defined as
\begin{equation}
    \alpha_{\mathcal{M}}(\lambda) \triangleq \max\limits_{\widetilde{\bm{\theta}}, d, d^{'}} \alpha_{\mathcal{M}}(\lambda; \bm{\theta}, d, d^{'})
\end{equation}
As we know from (13) and (14), we need the following conditions for differential privacy. Using the following Markov’s inequality
\begin{equation}
    \text{Pr}(X \geq a) \leq \frac{\mathbb{E}(X)}{a},
\end{equation}
for the LHS of (13), we have
\begin{align}
   \Pr\limits_{o \sim \mathcal{M}(d)}[c(o) \geq \epsilon] &= \Pr\limits_{o \sim \mathcal{M}(d)}[\exp(\lambda c(o)) \geq \exp(\lambda \epsilon)]  \leq \frac{\mathbb{E}_{o \sim \mathcal{M}(d)}[\exp(\lambda c(o))]}{\exp{ (\lambda \epsilon)}}.
\end{align}}

\blue{
For the RHS of (17), we have
\begin{align}
    \frac{\mathbb{E}_{o \sim \mathcal{M}(d)}[\exp(\lambda c(o))]}{\exp{ (\lambda \epsilon)}} = \frac{\exp( \log \mathbb{E}_{o \sim \mathcal{M}(d)}[\exp(\lambda c(o))])}{\exp{ (\lambda \epsilon)}} = \frac{\exp(\alpha_{\mathcal{M}}(\lambda; \bm{\theta}, d, d^{'}))}{\exp(\lambda\epsilon)} \\  \leq \frac{\exp(\max\limits_{\bm{W}, d, d^{'}} \alpha_{\mathcal{M}}(\lambda; \bm{\theta}, d, d^{'}))}{\exp(\lambda\epsilon)} = \frac{\exp[\alpha_{\mathcal{M}}(\lambda)]}{\exp(\lambda\epsilon)} = \exp(\alpha_{\mathcal{M}}(\lambda) - \lambda\epsilon).
\end{align}}

\blue{As presented in \cite{abadi2016deep}, $\alpha_{\mathcal{M}}(\lambda)$ have the composability as
\begin{equation}
    \alpha_{\mathcal{M}}(\lambda) \leq \sum_{t=1}^{T} \alpha_{\mathcal{M}_t}(\lambda)
\end{equation}}

\blue{For composability, we have the upper bound for $\exp(\alpha_{\mathcal{M}}(\lambda) - \lambda\epsilon)$ as
\begin{equation}
    \exp(\alpha_{\mathcal{M}}(\lambda) - \lambda \epsilon) \leq \exp\left(\sum\limits_{i=1}^{T}\alpha_{\mathcal{M}_{i}}(\lambda) - \lambda \epsilon \right).
\end{equation}
Then, we set $\delta$ as follows
\begin{equation}
    \delta = \min_{\lambda} \exp\left(\sum\limits_{i=1}^{T}\alpha_{\mathcal{M}_{i}}(\lambda) - \lambda \epsilon \right).
\end{equation}}

\blue{Using the following Lemma 6 from \cite{abadi2016deep}, we can bound $\alpha_{\mathcal{M}}(\lambda)$.
\begin{lemma} 
\label{lem:rdp}
Suppose that $f: D \rightarrow \mathbb{R}^d$ with $\|f(.)\|_2 \le 1$. Let $\sigma>1$ and let $J$ be a sample drawn i.i.d. from $[n]$ where each $i \in [n]$ is chosen independently with probability $q<\frac{1}{16\sigma}$. Then for any positive integer $\lambda\le\Bar{\rho}^2\sigma^{2}\ln\frac{1}{q\sigma}$, the mechanism $\mathcal{M}(d) = \sum_{i\in J}f(d_{i})+\mathcal{N}(0,\Bar{\rho}^2\sigma^{2}I)$ satisfies
\[\alpha_\mathcal{M}(\lambda) \leq \frac{q^2\lambda(\lambda+1)}{(1-q)\Bar{\rho}^2\sigma^2}+ O(q^3\lambda^3/\Bar{\rho}^{1.5}\sigma^3).\]
\end{lemma}
Together with the upper bound of $\alpha_{\mathcal{M}}(\lambda)$ above, we can verify the statement in (11) with the following derivation. 
From Lemma 6, $\alpha_{\mathcal{M}}(\lambda)$ is dominated by $c_2 \frac{q^2 \lambda^2}{\Bar{\rho}^2\sigma^2} $.  We have the following equation from (21).
\begin{equation}
    \delta \geq \min_{\lambda} \exp(c_2 \frac{T q^2 \lambda^2}{\Bar{\rho}^2 \sigma^2} - \lambda \epsilon)
\end{equation}
For $\lambda = \frac{- \epsilon \Bar{\rho}^2 \sigma^2}{2 c_2 q^2}$, we have the minimum value as $\frac{-\epsilon^2\Bar{\rho}^2\sigma^2}{4c_2Tq^2}$. Combine with (22), we have $\delta$ as
\begin{align}
    \delta &\geq \exp\left(\frac{-\epsilon^2\Bar{\rho}^2\sigma^2}{4c_2Tq^2}\right) \\
    \sigma &\geq 2c_2 \frac{q\sqrt{T\log(\frac{1}{\delta})}}{\Bar{\rho}\epsilon}
\end{align}}

\end{proof} 
\subsection{\blue{Additional Experiments}}
\label{sec:appendix_exp}
\blue{We provide the additional experiment results of the proposed methods on the financial domain.} We compare the financial tasks performance of the global model which is fine-tuned on the financial datasets. In our experiments, We assume that there are 5 clients in participating the whole training process. We use the above-mentioned models as our initial models. All the local models and global models follow the same size and use the original model weights in the initial. We trained each local model under the same set of parameters, such as learning rate, decay strategy, and training epochs. We set the compression rank $r$ to $512$ and $\epsilon$ value as $2$ when we vary the $\delta$ and $\delta$ as $1e-5$ when we vary the $\epsilon$. The origin row means that we train our model without differential private adaptations. The experimental data consistently demonstrate a clear trade-off between the privacy level (as adjusted by $\epsilon$ and $\delta$ values) and the model performance across various tasks. While all models show a decrease in performance with stricter privacy settings, the extent of this decrease varies, suggesting differences in how each model adapts to privacy constraints. 

One notable trend is the general decrease in performance across all models with stricter privacy settings (lower $\epsilon$
and $\delta$ values). For instance, GPT-2’s performance on MedQuAD drops from 69.2 at the original setting to $55.3$ and $49.1$
when $\epsilon$ is reduced to $2$ and $\delta$ to $1e-5$, respectively (Tables \ref{tab:epsilon_fin_gpt2} and \ref{tab:delta_fin_gpt2}). This trend indicates a trade-off between privacy and effectiveness.

In contrast, some models like Llama-7B show a more resilient performance under varying $\epsilon$ values. For example, its
performance on LiveQA only marginally decreases from $69.4$ to $66.1$ when $\epsilon$ is increased from the original to $10$ (Table
\ref{tab:epsilon_fin_llama}). This suggests that certain models might be better suited for privacy-sensitive applications.

Additionally, the impact of changing $\delta$ values appears to be more model-specific. Bert’s performance on MEDIQA-Ans
decreases significantly from $73.3$ in the original setting to $57.4$ when $\delta$ is reduced to $1e-6$ (Table \ref{tab:delta_fin_bert}), highlighting a potentially higher sensitivity to $\delta$ adjustments.

\input{tabs/delta_fin}

\input{tabs/epsilon_fin}

\begin{table}[]
\centering
\caption{The performance of the global model under different $\delta$ for GPT-2 in general and Medical tasks. }
\label{tab:delta_med_gpt}
\begin{tabular}{c|cccccc}
\toprule
$\delta $ & BoolQ  & PIQA   & WinoGrande& MedQuAD    & LiveQA & MEDIQA-Ans \\
\midrule
origin   &   61.9  &  53.8  &    48.3   &  69.2  &  39.5  &  55.1  \\
1e-02     & 60.2 & 50.7 & 46.6 & 69.1 & 32.5 & 51.1 \\
1e-03     & 54.7 & 49.5 & 38.7 & 59.8 & 36.1 & 46.1 \\
1e-04    & 61.6 & 47.6 & 35.0 & 68.5 & 30.9 & 50.3 \\
1e-05    & 51.5 & 46.3 & 46.1 & 55.3 & 27.0 & 49.1  \\
1e-06    & 53.4 & 42.9 & 40.6 & 69.1 & 30.5 & 41.4 \\
\bottomrule
\end{tabular}
\end{table}

\begin{table}[]
\centering
\caption{The performance of the global model under different $\delta$ for Bert in general and Medical tasks.}
\label{tab:delta_med_bert}
\begin{tabular}{c|cccccc}
\toprule
$\delta $ & BoolQ  & PIQA   & WinoGrande& MedQuAD    & LiveQA & MEDIQA-Ans \\
\midrule
origin   & 45.8  &  51.3 &   41.2  &  69.1  & 56.4   &  73.3 \\
1e-02    & 37.7 & 44.3 & 37.8 & 65.3 & 50.9 & 73.3 \\
1e-03     & 37.3 & 44.1 & 38.6 & 65.4 & 48.2 & 67.5 \\
1e-04    & 35.1 & 40.1 & 33.6 & 64.3 & 52.5 & 59.1 \\
1e-05    & 36.2 & 41.0 & 29.8 & 53.2 & 46.8 & 66.5 \\
1e-06    & 31.3 & 43.2 & 36.4 & 62.9 & 40.6 & 57.4 \\
\bottomrule
\end{tabular}
\end{table}

\begin{table}[]
\centering
\caption{The performance of the global model under different $\epsilon$ for GPT-2 in general and Medical tasks. }
\label{tab:epsilon_med_gpt2}
\begin{tabular}{c|cccccc}
\toprule
$\epsilon $ & BoolQ  & PIQA   & WinoGrande& MedQuAD    & LiveQA & MEDIQA-Ans \\
\midrule
origin   &   61.9  &  53.8  &    48.3   &  69.2  &  39.5  &  55.1 \\
2      & 51.5 & 46.3 & 46.1 & 55.3 & 27.0 & 49.1  \\
4      & 49.0 & 45.4 & 46.7 & 57.3 & 37.1 & 47.4 \\
6     & 58.4 & 41.5 & 45.6 & 63.5 & 38.9 & 52.9 \\
8      & 52.9 & 45.1 & 42.7 & 66.6 & 39.1 & 47.9 \\
10     & 58.9 & 53.1 & 44.7 & 64.1 & 35.9 & 51.8  \\
\bottomrule
\end{tabular}
\end{table}

\begin{table}[]
\centering
\caption{The performance of the global model under different $\epsilon$ for Bert in general and Medical tasks.}
\label{tab:epsilon_med_bert}
\begin{tabular}{c|cccccc}
\toprule
$\epsilon $ & BoolQ  & PIQA   & WinoGrande& MedQuAD    & LiveQA & MEDIQA-Ans \\
\midrule
origin   & 45.8  &  51.3 &   41.2  &  69.1  & 56.4   &  73.3 \\
2       & 36.2 & 41.0 & 29.8 & 53.2 & 46.8 & 66.5 \\
4       & 33.1 & 37.2 & 38.0 & 63.6 & 46.3 & 61.0 \\
6      & 40.1 & 40.2 & 33.1 & 68.8 & 47.6 & 60.9 \\ 
8      & 40.1 & 42.2 & 38.9 & 60.3 & 51.5 & 70.2 \\
10      & 42.2 & 45.9 & 37.9 & 68.9 & 51.6 & 69.2 \\
\bottomrule
\end{tabular}
\end{table}

\begin{table}[]
\centering
\caption{We show the communication overhead of GPT-2 in federated learning. We use the model parameter amount to denote the communication efficiency and compression ratio as an improvement indicator for different sets of rank $r$. The origin row means the results without our algorithm. For all the experiments, we train our local model under privacy parameter $\epsilon = 8$ and $\delta=10e^{-5}$. We also include the model performance on general tasks.}
\label{tab:compression_gpt2}
\begin{tabular}{c|cccccc|cc}
\toprule
rank $r$ & BoolQ & PIQA & WinoGrande & Communication overhead & Reduction ratio\\
\midrule
origin & 61.9 &  56.9 &  48.9   &   127.4M  &    \textbf{-}  \\
512    &  39.4 & 41.7 &   45.5   &   101.3M &    79.5\%    \\   
256    & 38.2 &  44.3 &  43.2   &   82.1M &   64.4\% \\ 
128    & 32.1 & 50.3 &  45.9   &  69.5M  & 54.6\%  \\
64     & 32.4 &  49.1 &  39.9   &  58.6M & 46.0\%  \\
32   & 33.5 & 46.3 &    39.4    &    58.0M &  45.5\%  \\
\bottomrule
\end{tabular}
\end{table}

\begin{table}[]
\centering
\caption{We show the communication overhead of Bert in federated learning. We use the model parameter amount to denote the communication efficiency and compression ratio as an improvement indicator for different sets of rank $r$. The origin row means the results without our algorithm. For all the experiments, we train our local model under privacy parameter $\epsilon = 8$ and $\delta=10e^{-5}$. We also include the model performance on general tasks.}
\label{tab:compression_bert}
\begin{tabular}{c|cccccc|cc}
\toprule
rank $r$ & BoolQ & PIQA & WinoGrande & Communication overhead & Reduction ratio\\
\midrule
origin & 49.8 &  55.5 &  60.2    &    110.8M  &  \textbf{-}  \\
512    & 35.5 & 44.7 &   41.3   &  83.7M  & 75.5\%  \\   
256    & 34.2 &  47.3 &  40.3   &   69.9M &   63.1\% \\ 
128    & 35.1 & 46.2 &  43.8  &  54.5M  & 49.2\%  \\
64     & 33.2 &  39.1 &  38.9  &  48.1M &  43.4\% \\
32     & 33.7 & 37.0 &    37.8    &    46.6M &  42.0\%  \\
\bottomrule
\end{tabular}
\end{table}

\end{document}

%% file: tabs/dataset_statics.tex
\begin{table}[t]
    \caption{Statistics of the training dataset}
    \label{tabs:statics}
    \begin{subtable}{.5\textwidth}
      \centering
        \caption{Statistics of the Medical Dialogue dataset. }
        \label{tabs:Medloge}
        \resizebox{!}{1.8cm}{
        \begin{tabular}{c|c}
        \hline
        Metric & Value \\
        \hline
        \# dialogues & 257,332 \\
        \# utterances & 514,664 \\
        \# tokens & 44,527,872 \\ \hline
        Avg. \# of utterances in a dialogue & 2 \\
        Max \# of utterances in a dialogue & 2 \\
        Min \# of utterances in a dialogue & 2 \\
        Avg. \# of tokens in an utterance & 86.5 \\
        Max \# of tokens in an utterance & 3,672 \\
        Min \# tokens in an utterance & 1 \\
        \hline
        \end{tabular}
        }
    \end{subtable}%
    \begin{subtable}{.5\textwidth}
      \centering
        \caption{Statistics of the Slimpajama. }
        \resizebox{!}{1.9cm}{
        \begin{tabular}{cc}
        \toprule
                     Data source & SlimPajama \\
        \midrule
                     Commoncrawl &      52.2\% \\
                              C4 &      26.7\% \\
                          GitHub &       5.2\% \\
                           Books &       4.2\% \\
                           ArXiv &       4.6\% \\
                       Wikipedia &       3.8\% \\
                   StackExchange &       3.3\% \\
        \bottomrule
\end{tabular}
        }
    \end{subtable} 
\end{table}

%% file: tabs/hyperparamter.tex
\begin{table}[]
\caption{hyper-parameter settings for training}
\begin{tabular}{cc}
\toprule
hyper-parameter & Value \\
\midrule
batch size $B$   &   $8$  \\
noise scale $\sigma $   &  $2$     \\
learning rate  $\gamma$  &   $5e^{-4}$    \\
clipping bound $C$  & $10$ \\   
\bottomrule
\end{tabular}
\label{tab:hyper}
\end{table}

%% file: tabs/epsilon_med.tex
\blue{
\begin{table}[]
\centering
\caption{The performance of the global model under different $\epsilon$ for ChatGLM-6B and Llama-7B in general and Medical tasks.}
\label{tab:epsilon_med}
\begin{tabular}{cccccccc}
\hline
Model                       & $\epsilon $ & BoolQ & PIQA & WinoGrande & MedQuAD & LiveQA & MEDIQA-Ans \\ \hline
\multirow{6}{*}{ChatGLM-6B} & origin      & 44.2  & 51.2 & 50.1       & 66.6    & 71.9   & 69.3       \\
                            & 2           & 34.0  & 47.9 & 34.9       & 66.4    & 67.3   & 57.8       \\
                            & 4           & 32.9  & 45.4 & 46.0       & 57.1    & 57.5   & 62.2       \\
                            & 6           & 37.8  & 42.2 & 47.5       & 60.6    & 58.9   & 65.8       \\
                            & 8           & 37.4  & 47.2 & 48.7       & 59.8    & 69.5   & 66.9       \\
                            & 10          & 38.8  & 47.9 & 43.8       & 65.9    & 69.8   & 67.7       \\ \hline
\multirow{6}{*}{Llama-7B}   & origin      & 71.3  & 66.2 & 65.5       & 73.1    & 69.4   & 65.5       \\
                            & 2           & 54.9  & 57.1 & 53.4       & 66.5    & 55.9   & 52.8       \\
                            & 4           & 60.6  & 59.5 & 55.5       & 66.6    & 67.3   & 53.2       \\
                            & 6           & 60.8  & 64.9 & 51.7       & 67.3    & 68.1   & 65.3       \\
                            & 8           & 66.3  & 64.1 & 60.4       & 66.2    & 60.5   & 62.1       \\
                            & 10          & 66.6  & 62.2 & 65.2       & 69.7    & 66.1   & 58.6       \\ \hline
\end{tabular}
\end{table}}

%% file: tabs/delta_med.tex
\begin{table}[]
\caption{The performance of the global model under different $\delta$ for ChatGLM-6B and Llama-7B in general and Medical tasks. The left column denotes the maximum value of $\delta$. }
\label{tab:delta_med}
\begin{tabular}{cccccccc}
\hline
Model                       & $\delta $ & BoolQ & PIQA & WinoGrande & MedQuAD & LiveQA & MEDIQA-Ans \\ \hline
\multirow{6}{*}{ChatGLM-6B} & origin    & 44.2  & 51.2 & 50.1       & 66.6    & 71.9   & 69.3       \\
                            & 1e-02     & 37.7  & 44.3 & 37.8       & 65.3    & 50.9   & 73.3       \\
                            & 1e-03     & 43.3  & 44.1 & 38.6       & 65.4    & 48.2   & 67.5       \\
                            & 1e-04     & 39.1  & 40.1 & 33.6       & 64.3    & 55.5   & 59.1       \\
                            & 1e-05     & 34.0  & 47.9 & 34.9       & 66.4    & 67.3   & 57.8       \\
                            & 1e-06     & 31.3  & 43.2 & 36.4       & 62.9    & 40.6   & 57.4       \\ \hline
\multirow{6}{*}{Llama-7B}   & origin    & 71.3  & 66.2 & 65.5       & 73.1    & 69.4   & 65.5       \\
                            & 1e-02     & 68.7  & 60.3 & 63.7       & 70.6    & 68.7   & 60.2       \\
                            & 1e-03     & 67.1  & 62.6 & 65.0       & 64.1    & 63.8   & 62.1       \\
                            & 1e-04     & 59.9  & 57.9 & 56.6       & 70.3    & 60.7   & 58.0       \\
                            & 1e-05     & 54.9  & 57.1 & 53.4       & 66.5    & 55.9   & 52.8       \\
                            & 1e-06     & 49.7  & 48.4 & 59.1       & 59.9    & 49.3   & 44.3       \\ \hline
\end{tabular}
\end{table}

%% file: tabs/compression.tex
\begin{table}[]
\centering
\caption{We show the communication overhead of ChatGLM-6B and Llama-7B in federated learning. We use the model parameter amount to denote the communication efficiency and compression ratio as an improvement indicator for different sets of rank $r$. The origin row means the results without our algorithm. For all the experiments, we train our local model under privacy parameter $\epsilon = 8$ and $\delta=10e^{-5}$. We also include the model performance on general tasks.}
\label{tab:compression}
\begin{tabular}{cccccccc}
\hline
Model                        & Method                & rank $r$ & BoolQ & PIQA & WinoGrande & Communication overhead & Reduction ratio \\ \hline
\multirow{10}{*}{Llama-7B} & Vanila FL \cite{mcmahan2017communication}            & -        & 76.3  & 79.7 & 70.0       & 6.7B                   & -               \\
                             & \blue{CMFL}~\cite{luping2019cmfl}                  & -        & \blue{71.4}   & \blue{77.3} & \blue{65.8}       & \blue{1.79B}                  &  \blue{26.8\%}     \\
                             & \blue{T-FedAvg}~\cite{xu2020ternary}              & -        & \blue{69.9}   & \blue{70.1} & \blue{67.0}       & \blue{1.18B}                  &  \blue{17.7\%}               \\ \cline{2-8} 
                             & \multirow{7}{*}{Ours} & 1024     & 78.8  & 83.1 & 71.1       & 2.43B                  & 36.27\%         \\
                             &                       & 512      & 76.4  & 80.4 & 69.9       & 1.35B                  & 20.15\%         \\
                             &                       & 256      & 70.6  & 74.1 & 68.3       & 0.93B                  & 13.88\%         \\
                             &                       & 128      & 50.3  & 78.5 & 69.7       & 0.65B                  & 9.70\%          \\
                             &                       & 64       & 39.1  & 68.5 & 67.8       & 0.49B                  & 7.31\%          \\
                             &                       & \blue{16}       & \blue{38.9}  & \blue{61.3} & \blue{45.7}       &  \blue{0.31B}                 & \blue{4.63\%}            \\
                             &                       & \blue{4}        & \blue{36.8}  &  \blue{50.7}    & \blue{32.3}       &  \blue{0.17B}                 &  \blue{2.53\%}             \\ \hline
\multirow{10}{*}{ChatGLM-6B}   & Vanila FL   \cite{mcmahan2017communication}           & -        & 69.5  & 76.3 & 71.3       & 6.2B                   & -               \\
                             & \blue{CMFL}~\cite{luping2019cmfl}                     & -        & \blue{55.2}  & \blue{64.9} & \blue{65.3}       & \blue{1.95B}                   &  \blue{31.4\%}         \\
                             & \blue{T-FedAvg}~\cite{xu2020ternary}                   & -        & \blue{53.9}  & \blue{68.6} & \blue{63.4}       & \blue{1.41B}                   &  \blue{22.9\%}          \\ \cline{2-8} 
                             & \multirow{7}{*}{Ours} & 1024     & 57.3  & 65.3 & 55.5       & 2.94B                  & 47.4\%          \\
                             &                       & 512      & 55.7  & 59.7 & 56.5       & 1.76B                  & 28.4\%          \\
                             &                       & 256      & 47.6  & 55.4 & 48.8       & 0.83B                  & 13.4\%          \\
                             &                       & 128      & 51.3  & 56.7 & 44.1       & 0.52B                  & 9.3\%           \\
                             &                       & 64       & 50.1  & 49.9 & 46.3       & 0.47B                  & 7.6\%           \\
                            &                       & \blue{16}       & \blue{41.3}  & \blue{47.7} & \blue{37.2}       & \blue{0.31B}                  & \blue{4.9\%}               \\
                             &                       & \blue{4}        & \blue{39.8}  & \blue{40.1} & \blue{36.9}       & \blue{0.15B}                  & \blue{2.4\%}               \\ \hline
\end{tabular}
\end{table}

%% file: tabs/delta_fin.tex
\begin{table}[]
\centering
\caption{The performance of the global model under different $\delta$ for GPT-2 in general and financial tasks. The left column denotes the maximum value of $\delta$.}
\label{tab:delta_fin_gpt2}
\begin{tabular}{c|cccccc}
\toprule
$\delta $ & BoolQ  & PIQA   & WinoGrande& FPB     & FiQA SA  & TFNS \\
\midrule
origin   &   41.1  &  39.9  &   37.5    &  59.8  &  73.8  &  66,0 \\
1e-02    & 36.4 & 39.6 & 35.7 & 57.1 & 71.4 & 57.0 \\
1e-03    & 34.7 & 35.4 & 29.4 & 51.1 & 63.3 & 52.2 \\
1e-04     & 33.1 & 28.6 & 27.8 & 42.1 & 57.5 & 45.1 \\
1e-05    & 25.5 & 20.5 & 22.2 & 39.1 & 42.9 & 34.3 \\
1e-06    & 23.2 & 19.0 & 20.3 & 28.1 & 52.2 & 34.0 \\
\bottomrule
\end{tabular}
\end{table}

\begin{table}[]
\centering
\caption{The performance of the global model under different $\delta$ for Bert in general and financial tasks. The left column denotes the maximum value of $\delta$.}
\label{tab:delta_fin_bert}
\begin{tabular}{c|cccccc}
\toprule
$\delta $ & BoolQ  & PIQA   & WinoGrande& FPB     & FiQA SA  & TFNS \\
\midrule
origin   &   45.3  &  47.5 &    51.3  &  65.5  &  60.1   &  57.3 \\
1e-02   & 41.8 & 41.7 & 46.7 & 59.6 & 59.1 & 55.6 \\
1e-03    & 37.9 & 39.7 & 44.9 & 50.6 & 55.6 & 50.9 \\ 
1e-04    & 37.7 & 36.8 & 39.2 & 48.4 & 51.4 & 47.9 \\
1e-05   & 21.7 & 29.9 & 38.8 & 43.7 & 47.5 & 44.1 \\
1e-06    & 20.3 & 28.7 & 30.2 & 37.3 & 45.3 & 40.3 \\
\bottomrule
\end{tabular}
\end{table}

\begin{table}[]
\centering
\caption{The performance of the global model under different $\delta$ for ChatGLM-6B in general and financial tasks. The left column denotes the maximum value of $\delta$. }
\label{tab:delta_fin_chatglm}
\begin{tabular}{c|cccccc}
\toprule
$\delta $ & BoolQ  & PIQA   & WinoGrande& FPB     & FiQA SA  & TFNS \\
\midrule
origin   &   61.5  &  69.3  &    63.8  &  55.5  &  74.3   &  70.1 \\
1e-02   & 55.5 & 68.7 & 61.1 & 50.7 & 69.4 & 62.2 \\
1e-03    & 53.6 & 65.3 & 54.9 & 49.1 & 64.5 & 56.2 \\
1e-04    & 50.7 & 62.3 & 51.7 & 47.4 & 58.9 & 47.6 \\
1e-05    & 44.9 & 58.3 & 48.6 & 29.9 & 48.1 & 42.5 \\
1e-06    & 38.5 & 55.2 & 47.4 & 30.6 & 51.5 & 38.0 \\
\bottomrule
\end{tabular}
\end{table}

\begin{table}[]
\centering
\caption{The performance of the global model under different $\delta$ for Llama-7B in general and financial tasks. The left column denotes the maximum value of $\delta$. \blue{Surprisingly, we observe that when $\delta=1e-02$, the performance is better. We hypothesize that the observed phenomenon may be attributed to the introduction of noise into the training process by a $\delta$ value of 1e-02. This noise could be beneficial, aiding the model in escaping local optima and consequently achieving improved results.}}
\label{tab:delta_fin_llama}
\begin{tabular}{c|cccccc}
\toprule
$\delta $ & BoolQ  & PIQA   & WinoGrande& FPB     & FiQA SA  & TFNS \\
\midrule
origin   &   76.3  &  79.7  &    70.0   &  59.1   &  80.0    &  69.6  \\
1e-02    &   88.1  &  80.6  &    69.3   &  80.3   &  81.9    &  67.3  \\
1e-03    &   81.8  &  78.2  &    57.6   &  64.1   &  75.0    &  65.4     \\ 
1e-04    &   75.4  &  69.3  &    59.1   &  49.1   &  76.1    &   70.3    \\ 
1e-05    &   59.3  &  71.5  &    61.6   &  38.5  &  48.3    &   43.2   \\
1e-06    &   69.5  &  56.6  &    63.5  &  35.2   &  51.5    &    28.9   \\
\bottomrule
\end{tabular}
\end{table}

%% file: tabs/epsilon_fin.tex
\begin{table}[]
\centering
\caption{The performance of the global model under different $\epsilon$ for GPT-2 in general and financial tasks. The left column denotes the maximum value of $\epsilon$.}
\label{tab:epsilon_fin_gpt2}
\begin{tabular}{c|cccccc}
\toprule
$\epsilon $ & BoolQ  & PIQA   & WinoGrande& FPB     & FiQA SA  & TFNS \\
\midrule
2      & 25.5 & 20.5 & 22.2 & 39.1 & 42.9 & 34.3 \\
4      & 26.8 & 19.5 & 29.3 & 40.8 & 53.9 & 35.2 \\
6      & 37.3 & 26.5 & 28.2 & 49.0 & 55.5 & 49.6 \\
8      & 33.9 & 27.5 & 31.1 & 53.2 & 56.9 & 54.7 \\
10     & 39.5 & 31.2 & 31.2 & 53.9 & 57.1 & 54.8 \\
origin & 41.1 & 39.9 & 37.5 & 59.8 & 73.8 & 66   \\
\bottomrule
\end{tabular}
\end{table}

\begin{table}[]
\centering
\caption{The performance of the global model under different $\epsilon$ for Bert in general and financial tasks. The left column denotes the maximum value of $\epsilon$.}
\label{tab:epsilon_fin_bert}
\begin{tabular}{c|cccccc}
\toprule
$\epsilon $ & BoolQ  & PIQA   & WinoGrande& FPB     & FiQA SA  & TFNS \\
\midrule
2       & 21.7 & 29.9 & 38.8 & 43.7 & 47.5 & 44.1 \\
4    & 24.5 & 33.5 & 39.9 & 45.2 & 47.3 & 44.5 \\
6     & 27.5 & 34.6 & 45.0 & 49.9 & 53.8 & 45.7 \\
8     & 31.0 & 36.2 & 46.7 & 50.4 & 54.9 & 51.9 \\
10     & 29.1 & 37.6 & 47.5 & 52.3 & 59.9 & 49.2  \\
origin  & 45.3 & 47.5 & 51.3 & 65.5 & 60.1 & 57.3 \\
\bottomrule
\end{tabular}
\end{table}

\begin{table}[]
\centering
\caption{The performance of the global model under different $\epsilon$ for ChatGLM-6B in general and financial tasks.. The left column denotes the maximum value of $\epsilon$.}
\label{tab:epsilon_fin_chatglm}
\begin{tabular}{c|cccccc}
\toprule
$\epsilon $ & BoolQ  & PIQA   & WinoGrande& FPB     & FiQA SA  & TFNS \\
\midrule
2      & 44.9 & 58.3 & 48.6 & 29.9 & 48.1 & 42.5 \\
4     & 43.2 & 58.3 & 48.3 & 40.2 & 57.9 & 54.8 \\
6     & 48.4 & 59.1 & 51.4 & 43.5 & 62.2 & 56.3 \\
8     & 52.5 & 60.9 & 58.0 & 45.6 & 68.1 & 63.2 \\
10   & 55.3 & 67.0 & 62.2 & 53.2 & 72.6 & 64.9 \\
origin & 61.5 & 69.3 & 63.8 & 55.5 & 74.3 & 70.1 \\
\bottomrule
\end{tabular}
\end{table}

\begin{table}[]
\centering
\caption{The performance of the global model under different $\epsilon$ for Llama-7B in general and financial tasks. The left column denotes the maximum value of $\epsilon$.}
\label{tab:epsilon_fin_llama}
\begin{tabular}{c|cccccc}
\toprule
$\epsilon $ & BoolQ  & PIQA   & WinoGrande& FPB     & FiQA SA  & TFNS \\
\midrule
2      &   59.6  &  71.5  &    61.6   &  38.5   &  48.3    &   43.2   \\
4      &   63.4  &  69.3  &    61.9   &  43.1   &  65.0    &  65.4     \\ 
6      &   71.9  &  70.3  &    59.1   &  49.1   &  76.1    &   65.3    \\ 
8      &   76.4  &  80.4  &    69.9   &  68.5  &   73.1    &   77.2   \\
10     &   87.9  &  80.1  &    71.1   &  69.9   &  81.5    &    82.9   \\
origin &   76.3  &  79.7  &    70.0   &  59.1   &  80.0    &  69.6  \\
\bottomrule
\end{tabular}
\end{table}

%% file: sample-manuscript.bbl
\begin{thebibliography}{60}
\providecommand{\natexlab}[1]{#1}
\providecommand{\url}[1]{\texttt{#1}}
\expandafter\ifx\csname urlstyle\endcsname\relax
  \providecommand{\doi}[1]{doi: #1}\else
  \providecommand{\doi}{doi: \begingroup \urlstyle{rm}\Url}\fi

\bibitem[Ouyang et~al.(2022)Ouyang, Wu, Jiang, Almeida, Wainwright, Mishkin, Zhang, Agarwal, Slama, Ray, et~al.]{ouyang2022training}
Long Ouyang, Jeffrey Wu, Xu~Jiang, Diogo Almeida, Carroll Wainwright, Pamela Mishkin, Chong Zhang, Sandhini Agarwal, Katarina Slama, Alex Ray, et~al.
\newblock Training language models to follow instructions with human feedback.
\newblock \emph{Advances in Neural Information Processing Systems}, 35:\penalty0 27730--27744, 2022.

\bibitem[OpenAI(2023)]{OpenAI2023GPT4TR}
OpenAI.
\newblock {GPT-4} technical report.
\newblock \emph{ArXiv}, abs/2303.08774, 2023.

\bibitem[Wu et~al.(2023)Wu, Irsoy, Lu, Dabravolski, Dredze, Gehrmann, Kambadur, Rosenberg, and Mann]{wu2023bloomberggpt}
Shijie Wu, Ozan Irsoy, Steven Lu, Vadim Dabravolski, Mark Dredze, Sebastian Gehrmann, Prabhanjan Kambadur, David Rosenberg, and Gideon Mann.
\newblock {BloombergGPT}: A large language model for finance.
\newblock \emph{arXiv preprint arXiv:2303.17564}, 2023.

\bibitem[Liu et~al.(2023{\natexlab{a}})Liu, Wang, Yang, and Zha]{liu2023fingpt}
Xiao-Yang Liu, Guoxuan Wang, Honyang Yang, and Daochen Zha.
\newblock {FinGPT}: Democratizing internet-scale data for financial large language models.
\newblock \emph{orkshop on Instruction Tuning and Instruction Following, NeurIPS}, 2023{\natexlab{a}}.

\bibitem[Yang et~al.(2023)Yang, Liu, and Wang]{yang2023fingpt}
Hongyang Yang, Xiao-Yang Liu, and Christina~Dan Wang.
\newblock {FinGPT}: Open-source financial large language models.
\newblock \emph{FinLLM at IJCAI}, 2023.

\bibitem[Liu et~al.(2023{\natexlab{b}})Liu, Xia, Yang, Gao, Zha, Zhu, Wang, Wang, and Guo]{liu2023dynamic}
Xiao-Yang Liu, Ziyi Xia, Hongyang Yang, Jiechao Gao, Daochen Zha, Ming Zhu, Christina~Dan Wang, Zhaoran Wang, and Jian Guo.
\newblock Dynamic datasets and market environments for financial reinforcement learning.
\newblock \emph{Machine Learning Journal, Springer Nature}, 2023{\natexlab{b}}.

\bibitem[Xie et~al.(2023)Xie, Han, Zhang, Lai, Peng, Lopez-Lira, and Huang]{xie2023pixiu}
Qianqian Xie, Weiguang Han, Xiao Zhang, Yanzhao Lai, Min Peng, Alejandro Lopez-Lira, and Jimin Huang.
\newblock Pixiu: A large language model, instruction data and evaluation benchmark for finance.
\newblock \emph{arXiv preprint arXiv:2306.05443}, 2023.

\bibitem[Singhal et~al.(2023)Singhal, Azizi, Tu, Mahdavi, Wei, Chung, Scales, Tanwani, Cole-Lewis, Pfohl, et~al.]{singhal2023large}
Karan Singhal, Shekoofeh Azizi, Tao Tu, S~Sara Mahdavi, Jason Wei, Hyung~Won Chung, Nathan Scales, Ajay Tanwani, Heather Cole-Lewis, Stephen Pfohl, et~al.
\newblock Large language models encode clinical knowledge.
\newblock \emph{Nature}, pages 1--9, 2023.

\bibitem[Nguyen(2023)]{nguyen2023brief}
Ha-Thanh Nguyen.
\newblock A brief report on lawgpt 1.0: A virtual legal assistant based on gpt-3.
\newblock \emph{arXiv preprint arXiv:2302.05729}, 2023.

\bibitem[Luo et~al.(2022)Luo, Sun, Xia, Qin, Zhang, Poon, and Liu]{luo2022biogpt}
Renqian Luo, Liai Sun, Yingce Xia, Tao Qin, Sheng Zhang, Hoifung Poon, and Tie-Yan Liu.
\newblock {BioGPT}: generative pre-trained transformer for biomedical text generation and mining.
\newblock \emph{Briefings in Bioinformatics}, 23\penalty0 (6):\penalty0 bbac409, 2022.

\bibitem[Zha et~al.(2023{\natexlab{a}})Zha, Bhat, Lai, Yang, Jiang, Zhong, and Hu]{zha2023data-centric-survey}
Daochen Zha, Zaid~Pervaiz Bhat, Kwei-Herng Lai, Fan Yang, Zhimeng Jiang, Shaochen Zhong, and Xia Hu.
\newblock Data-centric artificial intelligence: A survey.
\newblock \emph{arXiv preprint arXiv:2303.10158}, 2023{\natexlab{a}}.

\bibitem[Zha et~al.(2023{\natexlab{b}})Zha, Bhat, Lai, Yang, and Hu]{zha2023data-centric-perspectives}
Daochen Zha, Zaid~Pervaiz Bhat, Kwei-Herng Lai, Fan Yang, and Xia Hu.
\newblock {Data-centric AI}: Perspectives and challenges.
\newblock In \emph{SDM}, 2023{\natexlab{b}}.

\bibitem[Hu et~al.(2021)Hu, Shen, Wallis, Allen-Zhu, Li, Wang, Wang, and Chen]{hu2021lora}
Edward~J Hu, Yelong Shen, Phillip Wallis, Zeyuan Allen-Zhu, Yuanzhi Li, Shean Wang, Lu~Wang, and Weizhu Chen.
\newblock {LoRA}: Low-rank adaptation of large language models.
\newblock \emph{In International Conference on Learning Representations}, 2021.

\bibitem[Dodge et~al.(2020)Dodge, Ilharco, Schwartz, Farhadi, Hajishirzi, and Smith]{dodge2020fine}
Jesse Dodge, Gabriel Ilharco, Roy Schwartz, Ali Farhadi, Hannaneh Hajishirzi, and Noah Smith.
\newblock Fine-tuning pretrained language models: Weight initializations, data orders, and early stopping.
\newblock \emph{arXiv preprint arXiv:2002.06305}, 2020.

\bibitem[Yu et~al.(2021)Yu, Naik, Backurs, Gopi, Inan, Kamath, Kulkarni, Lee, Manoel, Wutschitz, et~al.]{yu2021differentially}
Da~Yu, Saurabh Naik, Arturs Backurs, Sivakanth Gopi, Huseyin~A Inan, Gautam Kamath, Janardhan Kulkarni, Yin~Tat Lee, Andre Manoel, Lukas Wutschitz, et~al.
\newblock Differentially private fine-tuning of language models.
\newblock \emph{International Conference on Learning Representations}, 2021.

\bibitem[Ding et~al.(2023)Ding, Qin, Yang, Wei, Yang, Su, Hu, Chen, Chan, Chen, et~al.]{ding2023parameter}
Ning Ding, Yujia Qin, Guang Yang, Fuchao Wei, Zonghan Yang, Yusheng Su, Shengding Hu, Yulin Chen, Chi-Min Chan, Weize Chen, et~al.
\newblock Parameter-efficient fine-tuning of large-scale pre-trained language models.
\newblock \emph{Nature Machine Intelligence}, 5\penalty0 (3):\penalty0 220--235, 2023.

\bibitem[Li et~al.(2023{\natexlab{a}})Li, Tan, and Liu]{li2023privacy}
Yansong Li, Zhixing Tan, and Yang Liu.
\newblock Privacy-preserving prompt tuning for large language model services.
\newblock \emph{arXiv preprint arXiv:2305.06212}, 2023{\natexlab{a}}.

\bibitem[Sebastian(2023)]{sebastian2023privacy}
Glorin Sebastian.
\newblock Privacy and data protection in chatgpt and other ai chatbots: Strategies for securing user information.
\newblock \emph{Available at SSRN 4454761}, 2023.

\bibitem[Li et~al.(2020{\natexlab{a}})Li, Fan, Tse, and Lin]{li2020review}
Li~Li, Yuxi Fan, Mike Tse, and Kuo-Yi Lin.
\newblock A review of applications in federated learning.
\newblock \emph{Computers \& Industrial Engineering}, 149:\penalty0 106854, 2020{\natexlab{a}}.

\bibitem[Zhang et~al.(2021)Zhang, Xie, Bai, Yu, Li, and Gao]{zhang2021survey}
Chen Zhang, Yu~Xie, Hang Bai, Bin Yu, Weihong Li, and Yuan Gao.
\newblock A survey on federated learning.
\newblock \emph{Knowledge-Based Systems}, 216:\penalty0 106775, 2021.

\bibitem[Carlini et~al.(2021)Carlini, Tramer, Wallace, Jagielski, Herbert-Voss, Lee, Roberts, Brown, Song, Erlingsson, et~al.]{carlini2021extracting}
Nicholas Carlini, Florian Tramer, Eric Wallace, Matthew Jagielski, Ariel Herbert-Voss, Katherine Lee, Adam Roberts, Tom Brown, Dawn Song, Ulfar Erlingsson, et~al.
\newblock Extracting training data from large language models.
\newblock In \emph{30th USENIX Security Symposium (USENIX Security 21)}, pages 2633--2650, 2021.

\bibitem[Kim et~al.(2023)Kim, Yun, Lee, Gubri, Yoon, and Oh]{kim2023propile}
Siwon Kim, Sangdoo Yun, Hwaran Lee, Martin Gubri, Sungroh Yoon, and Seong~Joon Oh.
\newblock Propile: Probing privacy leakage in large language models.
\newblock \emph{arXiv preprint arXiv:2307.01881}, 2023.

\bibitem[Li et~al.(2023{\natexlab{b}})Li, Guo, Fan, Xu, and Song]{li2023multi}
Haoran Li, Dadi Guo, Wei Fan, Mingshi Xu, and Yangqiu Song.
\newblock Multi-step jailbreaking privacy attacks on chatgpt.
\newblock \emph{arXiv preprint arXiv:2304.05197}, 2023{\natexlab{b}}.

\bibitem[Kone{\v{c}}n{\`y} et~al.(2016)Kone{\v{c}}n{\`y}, McMahan, Yu, Richt{\'a}rik, Suresh, and Bacon]{konevcny2016federated}
Jakub Kone{\v{c}}n{\`y}, H~Brendan McMahan, Felix~X Yu, Peter Richt{\'a}rik, Ananda~Theertha Suresh, and Dave Bacon.
\newblock Federated learning: Strategies for improving communication efficiency.
\newblock \emph{arXiv preprint arXiv:1610.05492}, 2016.

\bibitem[Hamer et~al.(2020)Hamer, Mohri, and Suresh]{hamer2020fedboost}
Jenny Hamer, Mehryar Mohri, and Ananda~Theertha Suresh.
\newblock Fedboost: A communication-efficient algorithm for federated learning.
\newblock In \emph{International Conference on Machine Learning}, pages 3973--3983. PMLR, 2020.

\bibitem[Lan et~al.(2023)Lan, Liu, Zhang, and Wang]{lan2023communication}
Guangchen Lan, Xiao-Yang Liu, Yijing Zhang, and Xiaodong Wang.
\newblock Communication-efficient federated learning for resource-constrained edge devices.
\newblock \emph{IEEE Transactions on Machine Learning in Communications and Networking}, 2023.

\bibitem[Carlini et~al.(2019)Carlini, Liu, Erlingsson, Kos, and Song]{carlini2019secret}
Nicholas Carlini, Chang Liu, {\'U}lfar Erlingsson, Jernej Kos, and Dawn Song.
\newblock The secret sharer: Evaluating and testing unintended memorization in neural networks.
\newblock In \emph{28th USENIX Security Symposium (USENIX Security 19)}, pages 267--284, 2019.

\bibitem[Shi et~al.(2021)Shi, Cui, Li, Jia, and Yu]{shi2021selective}
Weiyan Shi, Aiqi Cui, Evan Li, Ruoxi Jia, and Zhou Yu.
\newblock Selective differential privacy for language modeling.
\newblock \emph{arXiv preprint arXiv:2108.12944}, 2021.

\bibitem[Anil et~al.(2021)Anil, Ghazi, Gupta, Kumar, and Manurangsi]{anil2021large}
Rohan Anil, Badih Ghazi, Vineet Gupta, Ravi Kumar, and Pasin Manurangsi.
\newblock Large-scale differentially private bert.
\newblock \emph{arXiv preprint arXiv:2108.01624}, 2021.

\bibitem[Hoory et~al.(2021)Hoory, Feder, Tendler, Erell, Peled-Cohen, Laish, Nakhost, Stemmer, Benjamini, Hassidim, et~al.]{hoory2021learning}
Shlomo Hoory, Amir Feder, Avichai Tendler, Sofia Erell, Alon Peled-Cohen, Itay Laish, Hootan Nakhost, Uri Stemmer, Ayelet Benjamini, Avinatan Hassidim, et~al.
\newblock Learning and evaluating a differentially private pre-trained language model.
\newblock In \emph{Findings of the Association for Computational Linguistics: EMNLP 2021}, pages 1178--1189, 2021.

\bibitem[Li et~al.(2021)Li, Tramer, Liang, and Hashimoto]{li2021large}
Xuechen Li, Florian Tramer, Percy Liang, and Tatsunori Hashimoto.
\newblock Large language models can be strong differentially private learners.
\newblock \emph{arXiv preprint arXiv:2110.05679}, 2021.

\bibitem[Zhao et~al.(2023)Zhao, Zhou, Li, Tang, Wang, Hou, Min, Zhang, Zhang, Dong, et~al.]{zhao2023survey}
Wayne~Xin Zhao, Kun Zhou, Junyi Li, Tianyi Tang, Xiaolei Wang, Yupeng Hou, Yingqian Min, Beichen Zhang, Junjie Zhang, Zican Dong, et~al.
\newblock A survey of large language models.
\newblock \emph{arXiv preprint arXiv:2303.18223}, 2023.

\bibitem[Radford et~al.(2019)Radford, Wu, Child, Luan, Amodei, Sutskever, et~al.]{radford2019language}
Alec Radford, Jeffrey Wu, Rewon Child, David Luan, Dario Amodei, Ilya Sutskever, et~al.
\newblock Language models are unsupervised multitask learners.
\newblock \emph{OpenAI blog}, 1\penalty0 (8):\penalty0 9, 2019.

\bibitem[Brown et~al.(2020)Brown, Mann, Ryder, Subbiah, Kaplan, Dhariwal, Neelakantan, Shyam, Sastry, Askell, et~al.]{brown2020language}
Tom Brown, Benjamin Mann, Nick Ryder, Melanie Subbiah, Jared~D Kaplan, Prafulla Dhariwal, Arvind Neelakantan, Pranav Shyam, Girish Sastry, Amanda Askell, et~al.
\newblock Language models are few-shot learners.
\newblock \emph{Advances in neural information processing systems}, 33:\penalty0 1877--1901, 2020.

\bibitem[Black et~al.(2022)Black, Biderman, Hallahan, Anthony, Gao, Golding, He, Leahy, McDonell, Phang, et~al.]{black2022gpt}
Sid Black, Stella Biderman, Eric Hallahan, Quentin Anthony, Leo Gao, Laurence Golding, Horace He, Connor Leahy, Kyle McDonell, Jason Phang, et~al.
\newblock {GPT-NEOX-20B}: An open-source autoregressive language model.
\newblock \emph{arXiv preprint arXiv:2204.06745}, 2022.

\bibitem[Devlin et~al.(2018)Devlin, Chang, Lee, and Toutanova]{devlin2018bert}
Jacob Devlin, Ming-Wei Chang, Kenton Lee, and Kristina Toutanova.
\newblock Bert: Pre-training of deep bidirectional transformers for language understanding.
\newblock \emph{arXiv preprint arXiv:1810.04805}, 2018.

\bibitem[Chowdhery et~al.(2022)Chowdhery, Narang, Devlin, Bosma, Mishra, Roberts, Barham, Chung, Sutton, Gehrmann, et~al.]{chowdhery2022palm}
Aakanksha Chowdhery, Sharan Narang, Jacob Devlin, Maarten Bosma, Gaurav Mishra, Adam Roberts, Paul Barham, Hyung~Won Chung, Charles Sutton, Sebastian Gehrmann, et~al.
\newblock {PaLM}: Scaling language modeling with pathways.
\newblock \emph{arXiv preprint arXiv:2204.02311}, 2022.

\bibitem[Scao et~al.(2022)Scao, Fan, Akiki, Pavlick, Ili{\'c}, Hesslow, Castagn{\'e}, Luccioni, Yvon, Gall{\'e}, et~al.]{scao2022bloom}
Teven~Le Scao, Angela Fan, Christopher Akiki, Ellie Pavlick, Suzana Ili{\'c}, Daniel Hesslow, Roman Castagn{\'e}, Alexandra~Sasha Luccioni, Fran{\c{c}}ois Yvon, Matthias Gall{\'e}, et~al.
\newblock {BLOOM}: A {176B}-parameter open-access multilingual language model.
\newblock \emph{arXiv preprint arXiv:2211.05100}, 2022.

\bibitem[Zhang et~al.(2022)Zhang, Roller, Goyal, Artetxe, Chen, Chen, Dewan, Diab, Li, Lin, et~al.]{zhang2022opt}
Susan Zhang, Stephen Roller, Naman Goyal, Mikel Artetxe, Moya Chen, Shuohui Chen, Christopher Dewan, Mona Diab, Xian Li, Xi~Victoria Lin, et~al.
\newblock {OPT}: Open pre-trained transformer language models.
\newblock \emph{arXiv preprint arXiv:2205.01068}, 2022.

\bibitem[Touvron et~al.(2023)Touvron, Martin, Stone, Albert, Almahairi, Babaei, Bashlykov, Batra, Bhargava, Bhosale, et~al.]{touvron2023llama}
Hugo Touvron, Louis Martin, Kevin Stone, Peter Albert, Amjad Almahairi, Yasmine Babaei, Nikolay Bashlykov, Soumya Batra, Prajjwal Bhargava, Shruti Bhosale, et~al.
\newblock Llama 2: Open foundation and fine-tuned chat models.
\newblock \emph{arXiv preprint arXiv:2307.09288}, 2023.

\bibitem[Dettmers et~al.(2023)Dettmers, Pagnoni, Holtzman, and Zettlemoyer]{dettmers2023qlora}
Tim Dettmers, Artidoro Pagnoni, Ari Holtzman, and Luke Zettlemoyer.
\newblock {QLoRA}: Efficient finetuning of quantized llms.
\newblock \emph{arXiv preprint arXiv:2305.14314}, 2023.

\bibitem[Lester et~al.(2021)Lester, Al-Rfou, and Constant]{lester2021power}
Brian Lester, Rami Al-Rfou, and Noah Constant.
\newblock The power of scale for parameter-efficient prompt tuning.
\newblock \emph{arXiv preprint arXiv:2104.08691}, 2021.

\bibitem[Sung et~al.(2022)Sung, Cho, and Bansal]{sung2022lst}
Yi-Lin Sung, Jaemin Cho, and Mohit Bansal.
\newblock Lst: Ladder side-tuning for parameter and memory efficient transfer learning.
\newblock \emph{Advances in Neural Information Processing Systems}, 35:\penalty0 12991--13005, 2022.

\bibitem[Liu et~al.(2023{\natexlab{c}})Liu, Wang, Zhong, Xu, Zha, Tang, Jiang, Zhou, Chaudhary, Xu, et~al.]{liu2023winner}
Zirui Liu, Guanchu Wang, Shaochen Zhong, Zhaozhuo Xu, Daochen Zha, Ruixiang Tang, Zhimeng Jiang, Kaixiong Zhou, Vipin Chaudhary, Shuai Xu, et~al.
\newblock Winner-take-all column row sampling for memory efficient adaptation of language model.
\newblock \emph{arXiv preprint arXiv:2305.15265}, 2023{\natexlab{c}}.

\bibitem[Zaken et~al.(2021)Zaken, Ravfogel, and Goldberg]{zaken2021bitfit}
Elad~Ben Zaken, Shauli Ravfogel, and Yoav Goldberg.
\newblock Bitfit: Simple parameter-efficient fine-tuning for transformer-based masked language-models.
\newblock \emph{arXiv preprint arXiv:2106.10199}, 2021.

\bibitem[Karimi~Mahabadi et~al.(2021)Karimi~Mahabadi, Henderson, and Ruder]{karimi2021compacter}
Rabeeh Karimi~Mahabadi, James Henderson, and Sebastian Ruder.
\newblock Compacter: Efficient low-rank hypercomplex adapter layers.
\newblock \emph{Advances in Neural Information Processing Systems}, 34:\penalty0 1022--1035, 2021.

\bibitem[Houlsby et~al.(2019)Houlsby, Giurgiu, Jastrzebski, Morrone, De~Laroussilhe, Gesmundo, Attariyan, and Gelly]{houlsby2019parameter}
Neil Houlsby, Andrei Giurgiu, Stanislaw Jastrzebski, Bruna Morrone, Quentin De~Laroussilhe, Andrea Gesmundo, Mona Attariyan, and Sylvain Gelly.
\newblock Parameter-efficient transfer learning for nlp.
\newblock In \emph{International Conference on Machine Learning}, pages 2790--2799. PMLR, 2019.

\bibitem[McMahan et~al.(2017)McMahan, Moore, Ramage, Hampson, and y~Arcas]{mcmahan2017communication}
Brendan McMahan, Eider Moore, Daniel Ramage, Seth Hampson, and Blaise~Aguera y~Arcas.
\newblock Communication-efficient learning of deep networks from decentralized data.
\newblock In \emph{Artificial intelligence and statistics}, pages 1273--1282. PMLR, 2017.

\bibitem[Das et~al.(2022)Das, Acharya, Hashemi, Sanghavi, Dhillon, and Topcu]{das2022faster}
Rudrajit Das, Anish Acharya, Abolfazl Hashemi, Sujay Sanghavi, Inderjit~S Dhillon, and Ufuk Topcu.
\newblock Faster non-convex federated learning via global and local momentum.
\newblock In \emph{Uncertainty in Artificial Intelligence}, pages 496--506. PMLR, 2022.

\bibitem[Li et~al.(2020{\natexlab{b}})Li, Sahu, Zaheer, Sanjabi, Talwalkar, and Smith]{li2020federated}
Tian Li, Anit~Kumar Sahu, Manzil Zaheer, Maziar Sanjabi, Ameet Talwalkar, and Virginia Smith.
\newblock Federated optimization in heterogeneous networks.
\newblock \emph{Proceedings of Machine learning and systems}, 2:\penalty0 429--450, 2020{\natexlab{b}}.

\bibitem[Woodworth et~al.(2020)Woodworth, Patel, Stich, Dai, Bullins, Mcmahan, Shamir, and Srebro]{woodworth2020local}
Blake Woodworth, Kumar~Kshitij Patel, Sebastian Stich, Zhen Dai, Brian Bullins, Brendan Mcmahan, Ohad Shamir, and Nathan Srebro.
\newblock Is local sgd better than minibatch sgd?
\newblock In \emph{International Conference on Machine Learning}, pages 10334--10343. PMLR, 2020.

\bibitem[Karimireddy et~al.(2020)Karimireddy, Kale, Mohri, Reddi, Stich, and Suresh]{karimireddy2020scaffold}
Sai~Praneeth Karimireddy, Satyen Kale, Mehryar Mohri, Sashank Reddi, Sebastian Stich, and Ananda~Theertha Suresh.
\newblock Scaffold: Stochastic controlled averaging for federated learning.
\newblock In \emph{International conference on machine learning}, pages 5132--5143. PMLR, 2020.

\bibitem[Wei et~al.(2020)Wei, Li, Ding, Ma, Yang, Farokhi, Jin, Quek, and Poor]{wei2020federated}
Kang Wei, Jun Li, Ming Ding, Chuan Ma, Howard~H Yang, Farhad Farokhi, Shi Jin, Tony~QS Quek, and H~Vincent Poor.
\newblock Federated learning with differential privacy: Algorithms and performance analysis.
\newblock \emph{IEEE Transactions on Information Forensics and Security}, 15:\penalty0 3454--3469, 2020.

\bibitem[Abadi et~al.(2016)Abadi, Chu, Goodfellow, McMahan, Mironov, Talwar, and Zhang]{abadi2016deep}
Martin Abadi, Andy Chu, Ian Goodfellow, H~Brendan McMahan, Ilya Mironov, Kunal Talwar, and Li~Zhang.
\newblock Deep learning with differential privacy.
\newblock In \emph{Proceedings of the 2016 ACM SIGSAC conference on computer and communications security}, pages 308--318, 2016.

\bibitem[McSherry(2009)]{mcsherry2009privacy}
Frank~D McSherry.
\newblock Privacy integrated queries: an extensible platform for privacy-preserving data analysis.
\newblock In \emph{Proceedings of the 2009 ACM SIGMOD International Conference on Management of data}, pages 19--30, 2009.

\bibitem[Aghajanyan et~al.(2020)Aghajanyan, Zettlemoyer, and Gupta]{aghajanyan2020intrinsic}
Armen Aghajanyan, Luke Zettlemoyer, and Sonal Gupta.
\newblock Intrinsic dimensionality explains the effectiveness of language model fine-tuning.
\newblock \emph{arXiv preprint arXiv:2012.13255}, 2020.

\bibitem[Computer(2023)]{together2023redpajama}
Together Computer.
\newblock Redpajama: An open source recipe to reproduce llama training dataset, 2023.
\newblock URL \url{https://github.com/togethercomputer/RedPajama-Data}.

\bibitem[Zeng et~al.(2022)Zeng, Liu, Du, Wang, Lai, Ding, Yang, Xu, Zheng, Xia, et~al.]{zeng2022glm}
Aohan Zeng, Xiao Liu, Zhengxiao Du, Zihan Wang, Hanyu Lai, Ming Ding, Zhuoyi Yang, Yifan Xu, Wendi Zheng, Xiao Xia, et~al.
\newblock Glm-130b: An open bilingual pre-trained model.
\newblock \emph{arXiv preprint arXiv:2210.02414}, 2022.

\bibitem[Xu et~al.(2020)Xu, Du, Jin, He, and Cheng]{xu2020ternary}
Jinjin Xu, Wenli Du, Yaochu Jin, Wangli He, and Ran Cheng.
\newblock Ternary compression for communication-efficient federated learning.
\newblock \emph{IEEE Transactions on Neural Networks and Learning Systems}, 33\penalty0 (3):\penalty0 1162--1176, 2020.

\bibitem[Luping et~al.(2019)Luping, Wei, and Bo]{luping2019cmfl}
WANG Luping, WANG Wei, and LI~Bo.
\newblock Cmfl: Mitigating communication overhead for federated learning.
\newblock In \emph{2019 IEEE 39th international conference on distributed computing systems (ICDCS)}, pages 954--964. IEEE, 2019.

\end{thebibliography}
